\documentclass[pdflatex,sn-mathphys-num]{sn-jnl}

\usepackage{graphicx}%
\usepackage{multirow}%
\usepackage{amsmath,amssymb,amsfonts}%
\usepackage{amsthm}%
\usepackage{mathrsfs}%
\usepackage[title]{appendix}%
\usepackage{xcolor}%
\usepackage{textcomp}%
\usepackage{manyfoot}%
\usepackage{booktabs}%
\usepackage{algorithm}%
\usepackage{algorithmicx}%
\usepackage{algpseudocode}%
\usepackage{listings}%

\usepackage{subfig}
\usepackage{subcaption}

\usepackage{tikz}
\usetikzlibrary{shapes, shapes.geometric, arrows, calc}
\usetikzlibrary{decorations.pathreplacing} 

\definecolor{myred}{HTML}{CC0000}
\definecolor{myblue}{HTML}{A9C4EB}
\definecolor{myorange}{HTML}{FFCE9F}
\definecolor{mygreen}{HTML}{B9E0A5}
\definecolor{myviolet}{HTML}{CDA2BE}

\tikzstyle{data} = [
    trapezium,
    trapezium stretches=true, 
    trapezium left angle=70, 
    trapezium right angle=110, 
    minimum width=1cm, 
    minimum height=1cm, 
    text centered, 
    draw=black, 
    fill=myorange]
\tikzstyle{process} = [
    rectangle, 
    minimum width=1cm,
    minimum height=1cm, 
    text centered, 
    draw=black, 
    fill=myblue
]
\tikzstyle{code} = [
    trapezium,
    minimum width=1cm,
    minimum height=0.5cm,
    draw=black,
    trapezium left angle=70,
    trapezium right angle=70,
    shape border rotate=270,
    fill=mygreen
]
\tikzstyle{train} = [
    shape=signal,
    minimum width=1cm,
    minimum height=1cm,
    signal to=south,
    draw=black,
    align=center,
    signal to = east,
    fill = myviolet
]
\tikzstyle{darrow} = [thick,->,>=stealth] 
\tikzstyle{marrow} = [thick,<-,>=diamond,draw=myred] 
\tikzstyle{marrow1} = [thick,->,>=diamond,draw=myred] 
\tikzstyle{parrow} = [dotted,->,>=stealth] 

\theoremstyle{thmstyleone}%
\theoremstyle{thmstyletwo}%

\theoremstyle{thmstylethree}%

\begin{document}

\title[Article Title]{Hybrid twinning using PBDW and DeepONet for the effective state estimation and prediction on partially known systems}


\author*[1,2]{\fnm{Stiven} \sur{Massala}} 

\author[1, 2]{\fnm{Ludovic} \sur{Chamoin}} 

\author[2, 3, 4]{\fnm{Massimo Picca} \sur{Ciamarra}} 

\affil*[1]{\orgname{Universit\'e Paris-Saclay, CentraleSup\'elec, ENS Paris-Saclay, CNRS, LMPS - Laboratoire de M\'ecanique Paris-Saclay}, \orgaddress{\country{France}}}

\affil[2]{\orgname{CNRS@CREATE}, \orgaddress{ \country{Singapore}}}

\affil[3]{\orgdiv{Division of Physics and Applied Physics, School of Physical and Mathematical Sciences}, \orgname{
NTU Singapore}}

\affil[4]{\orgname{Consiglio Nazionale delle Ricerce, CNR-SPIN}, \orgaddress{ \country{Italy}}}


\abstract{The accurate estimation of the state of complex uncertain physical systems  requires reconciling theoretical models, with inherent imperfections, with noisy experimental data. In this work, we propose an effective hybrid approach that combines physics-based modeling with data-driven learning to enhance state estimation and further prediction. Our method builds upon the Parameterized Background Data-Weak (PBDW) framework, which naturally integrates a reduced-order representation of the best-available model with measurement data to account for both anticipated and unanticipated uncertainties. To address model discrepancies not captured by the reduced-order space, and learn the structure of model deviation, we incorporate a Deep Operator Network (DeepONet) constrained to be an orthogonal complement of the best-knowledge manifold. This ensures that the learned correction targets only the unknown components of model bias, preserving the interpretability and fidelity of the physical model. An optimal sensor placement strategy is also investigated to maximize information gained from measurements. We validate the proposed approach on a representative problem involving the Helmholtz equation under various sources of modeling error, including those arising from boundary conditions and source terms.}




\maketitle

\section{Introduction}
\label{section:introduction}
Effective modeling and simulation of physical phenomena need a proper union of theory and observation. It is in this spirit that data assimilation must be understood: this is a method by which experimental measurements are incorporated into physics-based modeling and simulation. Such physics-based models are rarely without imperfection, thus a central consideration in this process is whether the assimilation accounts for model bias. Classical bias-unaware approaches assume that the model is sufficiently accurate. They often begin by trusting the model veracity and subsequently calculate the bias, primarily relying on the calibration of parameters against presumed ground truth models. 
Contrarily, bias-aware data assimilation approaches are designed to contend with bias inherent in the assimilation process, in order to perform higher quality estimation from data \cite{inferring_novoa_2023}.  Such approaches fit into the paradigm of hybrid twins, in which physics-based simulation models are informed and enriched from the status of the studied physical system, which is probed via sensors. An hybrid twin relies on a deviation model by correcting a model bias or learning the model ignorance, and integrating this deviation model within the physics-based simulations. This class of approaches offers a meaningful trade-off between data-driven and physics-driven modeling strategies. Indeed, purely data-driven models, while flexible, tend to suffer from high variance and require extensive datasets. Conversely, purely model-based simulations, though stable and data-efficient, are prone to model bias and prohibitive computational cost when high fidelity is targeted. Hybrid twins, as introduced in \cite{virtual_chinesta_2020}, leverage the best of both worlds.\\

In this paper, we deal with one such hybrid twinning method, the Parameterized Background Data-Weak (PBDW) which is a non-intrusive, reduced basis, and naturally bias-aware data assimilation framework for physical systems modeled by parametrized Partial Differential Equations (PDE) \cite{parameterizedbackground}. The method approximates the system state by combining:
\begin{itemize}
    \item A background component, derived from a reduced-order model over a parameter space \cite{PrudHomme2002Reliable};
    \item An update component, which corrects the unmodeled physics through the Riesz representation of observation functionals.
\end{itemize}

These components are constructed simultaneously in orthogonal subspaces, which is an attractive feature of PBDW that ensures that the correction is optimal and non redundant. Moreover, the method includes an optimal sensor placement strategy that guarantees computational efficiency. Initially developed for steady-state problems, PBDW has been adapted to handle noisy observations \cite{adaptive_taddei_2017,adaptive_maday_2019}, extended to time-dependent systems using the POD-Greedy approach \cite{Benaceur2019ATP}, and further refined to enable simultaneous estimation of both system state and parameters as unbiased noisy data are acquired sequentially \cite{realtime_haik_2023}.   The work in \cite{Galarce2024} extends the classical approach to embrace the presence of biased measurement noise. The results on the stability and accuracy of the PBDW method have been established in \cite{Binev2017}. Building on this foundation, PBDW has been effectively applied to a wide range of problems, including nuclear reactor simulations \cite{Argaud2018}, thermal analysis \cite{Benaceur2021}, and advanced medical imaging techniques \cite{Galarce2021a,Galarce2021b}.\\

Complementarily, recent development in Deep Learning models \cite{DL,LeCun2015DeepLearning,Schmidhuber2015DeepLearning} have shown success in approximating solutions of various physics models. Physics-informed neural networks (PINNS) \cite{raissi2019physics}  were of the first NN-based approaches to be introduced as effective surrogate models when solving forward or inverse problems modeled by PDEs. Being fully differentiable thanks to automatic differentiation \cite{Paszke2017AutomaticDI,Bischof2000ComputingDO,Merrienboer2017Tangent}, these modeling approaches introduce a physics-guided loss as a regularizer. So far, several frameworks
have been proposed to take into account physical properties in the architecture, such as thermodynamics constraints  \cite{NN-mCRE,anisotropic,mechanics-informed,NN-euclid,modular-ML,VONNs,structure-preserving-NN}. 

The generalization of the universal approximation theorem from functions to operators \cite{Arbitrary-Depth} also led to the development of neural operators, designed not to approximate single functions, but mappings between functions. Among them, the Deep Operator Network (DeepONet) \cite{lulu} employs two distinct components: one to represent the input function via sampled sensor data (branch network), and another to encode the output locations (trunk network). Another approach is the Fourier Neural Operator (FNO) \cite{FourierNeuralOperator}, which learns the solution operator by parametrizing the kernel in the Fourier domain. These operator architectures achieve remarkable accuracy in solving forward and inverse problems associated with partial differential equations, suggesting a promising path toward data-driven modeling of physical phenomena.\\

This rapid expansion of scientific machine learning tools has induced some interest in integrating data-driven methods within data assimilation techniques~\cite{Bocquet2020}. 
Such approaches are especially useful for correcting model errors. 
For example,~\cite{Bonavita2020,Farchi2021,NN-mCRE} use neural networks inside variational frameworks. 
Other works~\cite{activelearning,learncorrect} propose digital twins that mix machine learning and computer vision, along with augmented reality. 
More recently, several studies have focused on explicitly learning and modeling structural or physical discrepancies in dynamical systems, such as the frameworks proposed in~\cite{Wu2023,Ebers2023,Wu2024}, 
while~\cite{inferring_novoa_2023} introduced a method that can accurately estimate model bias.

In practice, unmodeled physical phenomena (model bias)  represent 'unknown unknowns' that are challenging to detect and quantify \cite{inferring_novoa_2023}. Deep learning regression thus offers a powerful framework to identify and learn this model ignorance. For example, machine learning techniques have been proposed to approximate the solution error in parameterized equations \cite{Trehan2017ErrorModeling}, and later extended to dynamical systems \cite{Trehan2022MachineLearning}. It is therefore appropriate to learn the bias structure with a neural operator, as it enables efficient inference of physical quantities of interest.\\ 

In this context, the present study proposes a coupling between DeepONet \cite{lulu} and the Parametrized-Background Data-Weak (PBDW) method~\cite{parameterizedbackground}. A modified DeepONet architecture is introduced as a deviatoric model, designed to approximate localized unmodeled physics by means of correction coefficients. In this approach, local quantities are mapped from discrete sensor observations to a continuous field via the Riesz representers of the observation functionals. To preserve the orthogonality, which is central in the PBDW framework, the DeepONet correction is confined to the orthogonal complement of the background space, thereby enforcing that only unmodeled physics is learnt.
Furthermore, the spatial discretization of the output (trunk) network is optimized such that trunk nodes are interpreted as sensor locations, selected via the greedy procedure introduced in \cite{stability}. \\

The proposed hybrid-AI framework provides these contributions: \\

1. Reformulation of the PBDW correction mechanism where the bias is approximated by a deep operator network (DeepONet). 

2. Proposition of a DeepONet that is informed by observables  \cite{Simultaneous}, whose number of trainable parameters scales with the number of experimental observations.

3. DeepONet correction designed to capture the most useful experimental data, with an optimally discretized trunk net in which  each node corresponds to a sensor optimally placed. Optimality is here understood in the sense of stability maximization of the background and minimization of the approximation error \cite{stability}. 

4. A DeepONet correction model restricted to the orthogonal complement of the reduced-order background, enforced via a Gram-Schmidt orthonormalization of the reduced basis. This ensures that the learned correction strictly targets the unmodeled components of the physical system.\\

As shown in the numerical results, the proposed architecture offers significant advantages over purely data-driven approaches, in particular: improved robustness to overfitting, enhanced interpretability, resilience outside the training regime, and compatibility with real-time estimation and control. It thus represents a promising direction for bias-aware data assimilation in the presence of model uncertainty.\\

The paper is organized as follows: In Section 2 we present the PBDW and the algebraic formulation of the assimilation task associated with noisy measurements. Section 3 introduces DeepONet and its coupling with PBDW. An analysis of complexity is presented in Section 4. Numerical experiments are conducted in Section 5 to assess and validate the performance of the proposed approach. Eventually, conclusions and prospects are drawn in Section 6.

\section{ Parameterized-Background Data-Weak (PBDW)}

This section is an introduction to the PBDW method which was originally developed in a steady framework in \cite{parameterizedbackground}. We introduce the Hilbert space $\mathcal{U}$ over a spatial domain $\Omega$ endowed with the inner product $(\cdot,\cdot)$ and the induced norm $\|\cdot \|$. We denote by $\mu \in D$ the parametric configuration of the physical system, where $D \subset \mathbb{R}^d$ is the non-empty set of all parameters of interest.

\subsection{Formulation of PBDW } 

The knowledge embedded in the parameterized PDE $\mathcal{P}^{bk}$ reflects a physical model based on certain assumptions and is used to compute the background state $u^{bk}(\mu) \in \mathcal{U}$. This leads to the definition of the background manifold $\mathcal{M}^{bk} = \{ u^{bk}(\mu) \mid \mu \in D \}$, which consists of solution snapshots of $\mathcal{P}^{bk}$ across $D$. Under the assumption that $\mathcal{M}^{bk}$ does not perfectly capture the true physics, the true state $u_{\text{true}}(\mu)$ lies outside the manifold $\mathcal{M}^{bk}$.\\

Experimental data $\{y_{m}\}$ are collected with observation functionals $l_m$ that describe the transfer function of the sensors, and are assumed to be linear on $\mathcal{U}$ :
\begin{equation}
y_m=l_m\left(u_{\text {true }}\right)+\epsilon_m \quad 1 \leq m \leq M
\end{equation}
with $l_m$ being e.g. a local average of the state in the neighborhood of the location $x_m \in \Omega$ and $\{\epsilon_m \}$ some observational noise. We define the set of sensor positions $\mathcal{X}=\left\{x_m\right\}_{m=1}^M$.\\

Given the model $\mathcal{P}^{bk}$ and $M$ experimental data, the PBDW method estimates $u_{\text {true }}$ as a sum of contributions :
\begin{equation}
u_{N, M}=z_{N, M}+\eta_{N, M}
\end{equation}

The first (background) contribution $z_{N, M} \in \mathcal{Z}_N=\operatorname{span}\left\{\zeta_n\right\}_{n=1}^N \subset \mathcal{U}$ is searched in a linear reduced-order space representing $\mathcal{M}^{b k}$ (the reduction being achieved using the Reduced Basis Method~\cite{Rozza2007}, the Generalized Empirical Interpolation Method~\cite{Barrault2004}, or nonlinear approaches tailored to specific inverse problems~\cite{Cohen2022}). The fact that the true state $u_{\text{true}}$ does not live in $\mathcal{M}^{b k}$ motivates the second contribution $\eta_{N, M} \in \mathcal{U}_M = \operatorname{span}\left\{q_m\right\}_{m=1}^M $ called update. The update contribution accommodates unanticipated uncertainty, and $q_m$ are the Riesz representers  of $\left\{l_m\right\}_{m=1}^M$ defined such that $ \; (q_m,u) = l_m(u) \; \forall v \in \mathcal{U}, m=1,\dots,M $.\\ 

The philosophy is to look for the smallest correction $\eta_{N, M}$ to the best-knowledge  estimate $z_{N, M}$, subject to the observation constraints $\ell(\eta_{N, M}+z_{N, M})=\ell\left(u^{true}\right) + \epsilon_{m}$.  A Tikhonov regularization was introduced in \cite{adaptive_maday_2019} to account for measurement noise, leading to the following minimization problem:\\ 

Find $\left(z_{N, M}^{\xi}, \eta_{N, M}^{\xi}\right) \in \mathcal{Z}_N \times \mathcal{U}_M$ such that:

\begin{equation}
\left(z_{N, M}^{\xi}, \eta_{N, M}^{\xi}\right)=\underset{\substack{z \in \mathcal{Z}_N \\ \eta \in \mathcal{U}_M}}{\arg \inf }\left(\xi\|\eta\|^2+\frac{1}{M} \sum_{m=1}^M\left\|l_m(z+\eta)-y_m\right\|_2^2\right)
\label{eq:pbdw_regul}
\end{equation}
with $\|\cdot\|_2$ the Euclidean $L^2$-norm in $\mathbb{R}^M$, and $\xi$ the regularization weight which allows tuning the approximation between $\mathcal{Z}_N$ and $\mathcal{U}_M$. It is a regularized nonlinear least-square problem; its formulation reduces to the well-known 3D-VAR formulation when $\mathcal{M}^{b k}$ reduces to a singleton. \\

We now define matrices $\mathbb{B} \in \mathbb{R}^{M \times N}$ and $\mathbb{A} \in \mathbb{R}^{M \times M}$ such that:
\begin{equation}
B_{m,n} = l_m(\zeta_n) \quad ; A_{m,m'} = l_m(q_{m'})
\end{equation}

Then, we define vector-valued functions of coordinates ${\mathbf{z}^{\star} } \in \mathbb{R}^N$ and ${\boldsymbol{\eta}^{\star}} \in \mathbb{R}^M$ such that:

\begin{equation}
u(x)_{N, M} = \sum_{n=1}^N(\mathbf{z}^{\star})_n \zeta_n(x) + \sum_{m=1}^M (\boldsymbol{\eta}^{\star})_m q_m(x)
\end{equation}
By rescaling $\tilde{\xi} = M \xi$, the algebraic formulation is the following discrete optimization problem:
\begin{equation}
(\mathbf{z}^{\star},\boldsymbol{\eta}^{\star}) = \underset{\substack{\mathbf{z} \in \mathbb{R}^N \\ \boldsymbol{\eta} \in \mathbb{R}^M}}{\arg \min } \\ \left(\tilde{\xi} \boldsymbol{\eta}^T \mathbb{A} \boldsymbol{\eta} + \| \mathbb{A} \boldsymbol{\eta} + \mathbb{B}\mathbf{z}  - \mathbf{y}\|^2_2\right)
\end{equation}
 
Since this problem is quadratic in $\boldsymbol{\eta}$, we have the following relationship between $\boldsymbol{\eta}^{\star}$ and $\mathbf{z}^{\star}$:
\begin{equation}
\boldsymbol{\eta}^{\star} = \mathbb{W}_{\xi}(\mathbf{y}-\mathbb{L}\mathbf{z}^{\star}) \quad ; \quad \mathbb{W}_{\xi} = (\tilde{\xi} \mathbb{I}_d+\mathbb{A})^{-1}
\end{equation}
which can be restated as $\mathbb{A} \boldsymbol{\eta}^{\star} + \mathbb{B}\mathbf{z}^{\star} -\mathbf{y} = - \tilde{\xi} \boldsymbol{\eta}^{\star}$. By substituting, we then find that:
\begin{equation}
\tilde{\xi} \boldsymbol{\eta}^{\star ^T} \mathbb{A} \boldsymbol{\eta}^{\star} +  \|\mathbb{A} \boldsymbol{\eta}^{\star} + \mathbb{B}\mathbf{z}^{\star} -\mathbf{y} \|^2_2 = \xi(\mathbf{y}-\mathbb{B}\mathbf{z}^{\star})^T \mathbb{W}_{\xi} \mathbb{W}_{\xi}^{-1}(\mathbf{y}-\mathbb{B}\mathbf{z}^{\star})
\end{equation}
which implies:
\begin{equation}
\mathbf{z}^{\star} = \underset{\substack{\mathbf{z} \in \mathbb{R}^N}}{\arg \min } \\  \| \mathbf{y}-\mathbf{B}\mathbf{z}^{\star}\|_{\mathbf{W}_{\xi}}
\end{equation}

The problem can be decomposed into an $N$-dimensional constrained optimization problem followed by an $M$-dimensional linear problem.  This decomposition makes it suitable to reconstruct the solution in an hybrid way;  hybrid in the sense that the background is the solution of an $N$-dimensional constrained optimization problem defined from physics, while the update is approximated from data. A novel strategy that represents the update by means of a deep operator network (DeepONet) will be detailed in the next chapter.\\

The problem above is well-posed under the satisfaction of the condition : \hspace{0.15cm}
\begin{equation}
\beta_{N, M}=\inf _{z \in \mathcal{Z}_N} \sup _{q \in \mathcal{U}_M} \frac{(z, q)}{\|z\|\|q\|} > 0 \\
\label{eq:beta_equation}
\end{equation}
$\beta_{N, M}$ is geometrically interpreted as the angle between the reduced space $\mathcal{Z}_N$ and the observation space $\mathcal{U}_M$. It measures how well the elements in $\mathcal{Z}_N$ are observed by $\mathcal{U}_M$. In the context of noise-free data and considering a non-regularized PBDW statement$(\xi=0)$, an a priori error estimation was developed in \cite{parameterizedbackground} using $\beta_{N, M}$. The analysis in \cite{parameterizedbackground} gives a recovery of the true state with the bound:

\begin{equation}
    \left\|u_{\text {true }}-u_{N, M}\right\| \leq\left(1+\frac{1}{\beta_{N, M}}\right) \inf _{q \in \mathcal{U}_M \cap \mathcal{Z}_{N}^{\perp}}\left\|\Pi_{\mathcal{Z}_{N}^{\perp}} u_{\text {true }}-q\right\|
\end{equation}

 This bound shows how the quality (approximation error) of the reconstruction strongly depends on its stability ($\beta_{N, M}$). Since $\beta_{N, M}$ is a non-increasing function of  $N$ (the dimension of $\mathcal{Z}_N$) and a non-decreasing function of  $M$ (the dimension of $\mathcal{U}_M$), the spaces $\mathcal{U}_M$ and $\mathcal{Z}_N$ are chosen such that $\beta_{N, M}$ is maximized.\\

Hence, to have the best recovery (minimize the approximation error), we want to:\\
\begin{itemize}
    \item Maximize the inf-sup constant $\beta_{N, M}$;
    \item Minimize the background best-fit error $\epsilon_N^{b k}=\left\|\Pi_{\mathcal{Z}_{N}} u_{\text {true }}\right\|$ ;
    \item Minimize the update best-fit error $\inf _{q \in \mathcal{U}_M \cap \mathcal{Z}_{N}^{\perp}}\left\|\Pi_{\mathcal{Z}_{N}} u_{\text {true }}-q\right\|$.  This error is given by the projection of the portion of the true state not approximated by the background space. \\
\end{itemize} 

The condition $\beta_{N, M} > 0$ implies $M\geq N$. This means that in practical engineering systems, the number of experimental observations $M$ is bounded by the value of $N$ minimizing $\epsilon_N^{b k}$. Hence, when $M$ is close to $N$, we need to make the experiments smart by choosing the sensors optimally.




\subsection{ Orthogonality property of PBDW } 

Let us come back to the algebraic formulation introduced in the previous section.
We now know that any element $z \in \mathcal{Z}_N$ is expressed as $z=\sum_{n=1}^N z_n \zeta_n$; any element $\eta \in \mathcal{U}_M$ is expressed as $\eta=\sum_{m=1}^M \eta_m q_m$; the reconstructed state is given by $u_{N, M}^*= \sum_{n=1}^N z_n^* \zeta_n + \sum_{m=1}^M \eta_m^* q_m$.\\

The algebraic formulation of the regularized PBDW reads:

\begin{equation}
\left[\begin{array}{cc}
\xi M \mathbb{I}_M+\mathbb{A}   &  
 \quad \mathbb{B} \\
\mathbb{B}^T & 0
\end{array}\right]\left[\begin{array}{l}
\boldsymbol{\eta}^{\star} \\
\mathbf{z}^{\star}
\end{array}\right]=\left[\begin{array}{l}
\boldsymbol{y} \\
\mathbf{0}
\end{array}\right]
\label{eq:saddle}
\end{equation}

The pair $\left(\boldsymbol{\eta}^{\star}, \boldsymbol{z}^{\star}\right) \in \mathbb{R}^M \times \mathbb{R}^N$ is solution of the system:
\begin{equation}
\centering
\begin{cases}(\xi M \mathbb{I}+\mathbb{A}) \boldsymbol{\eta}^{\star}+\mathbb{B} \mathbf{z}^{\star} & =\boldsymbol{y} \\ \mathbb{B}^T \boldsymbol{\eta}^{\star} & =\mathbf{0} .\end{cases}
\label{eq:formulation}
\end{equation}

Given the matrix $B_{m,n} = l_m(\zeta_n)$, and the background space $\mathcal{Z}_N=\operatorname{span}\left\{\zeta_n\right\}_{n=1}^N$, the equation $(\ref{eq:formulation})_{2}$ gives the following orthogonality property :
\begin{equation}
\mathbb{B}^T \boldsymbol{\eta}^{\star} = 0 \iff   \left(\eta^*, z\right)=0 \quad \forall z \in \mathcal{Z}_N \iff \eta_{N, M}^* \in \mathcal{U}_M \cap \mathcal{Z}_N^{\perp}\\
\end{equation}
meaning that $\eta_{N, M}^*$ only accommodates the part of the projection onto $\mathcal{U}_M$ which cannot be absorbed by $z_{N, M}^* \in \mathcal{Z}_N$, that is the part that lies in $\mathcal{Z}_N^{\perp}$ \cite{parameterizedbackground}. This is an important property that we should preserve when approximating $\eta_{N, M}$ by a deep neural operator in the following.

\section{ Coupling PBDW with DeepONet }

    
    

    
    

    
    


Deep Operator Networks (DeepONets) are a generalization of standard deep learning techniques that learn mappings between function spaces rather than finite-dimensional vector spaces. A DeepONet consists of two independent neural networks with general architectures: the branch network and the trunk network.
The branch network $\{b_k\}_{k=1}^p$ extracts latent representations of input functions sampled at sensor locations $\{x_m\}_{m=1}^M$, processing the discretized input $\{v(x_m)\}_{m=1}^M$. The trunk network $\{t_k\}_{k=1}^p$ extracts latent representations of spatial coordinates $y$ at which the output function is to be evaluated. These networks are coupled through a basis-coefficient structure, yielding the approximation:
\begin{equation}
G(v)(y) \approx \sum_{k=1}^p b_k(\{v(x_m)\}_{m=1}^M) \cdot t_k(y)
\end{equation}

DeepONets can be enhanced when a priori knowledge of the system is available, by replacing the trunk network with a set of basis functions derived from a background knowledge. In POD-DeepONet for instance, the trunk network is substituted with Proper Orthogonal Decomposition (POD) basis functions computed from training data. Similarly, SVD-DeepONet \cite{ventu} incorporates low-rank representations through Singular Value Decomposition. Physics-informed constraints can also be integrated into the architecture. For instance, \cite{jnini2024physicsconstrained} imposed divergence-free constraints derived from the continuity equation for fluid mechanics applications. In this work, we present a PBDW-based framework that employs neural operators to recover the unmodeled physics, specifically the discrepancies that physics-based models do not capture, thereby allowing for accurate representation of the true state $u$.

We recall the PBDW decomposition:
\begin{equation}
u_{N, M}=z_{N, M}+\eta_{N, M}=\Pi_{\mathcal{Z}_{N}} u^{\text {true }} +  \Pi_{\mathcal{Z}_{N}^{\perp}} u^{\text {true }}
\end{equation}

We propose a hybrid approach that combines the classical PBDW framework with operator learning. The background component $z_{N, M} \in \mathcal{Z}_{N}$, which encodes the underlying physics, is computed through the standard PBDW saddle-point formulation. However, the update term $\eta_{N, M} \in \mathcal{U}_{M}$ is now parameterized using a DeepONet architecture. Consider a physical system governed by a parametric PDE with input function $v$ representing the forcing term or inhomogeneous source. By selecting $K$ distinct forcing functions $v^{(i)}$, we generate a training dataset $\mathcal{T}$ by solving $K$ PBDW problems with corresponding PBDW update term $\eta^{(i)}$:
\begin{equation}
\mathcal{T}=\left\{\left(v^{(i)}, \eta^{(i)}\right)\right\}_{i=1}^{K}
\end{equation}

Given the a priori subspace $\mathcal{Z}_{N}$ and training set $\mathcal{T}$, we train a neural operator $\mathcal{G}_{\text{update}}$ to predict the update coefficients. It reads:
\begin{equation}
\mathcal{G}_{update} : \begin{cases} v \in\mathcal{V}, v \mapsto \eta_{\theta}(v)(x_m) \\ \text { Such that } \sum_{m=1}^M q_m \cdot \eta_{\theta}(v)(x_m)    \in \mathcal{U} \cap \mathcal{Z}_N^{\perp} & 
\end{cases}
\label{eq:map}
\end{equation}
For a new forcing function $v$, the neural operator outputs discrete values $\boldsymbol{\eta}_{\theta} = [\eta_{\theta}(x_1), \ldots, \eta_{\theta}(x_M)]^T \in \mathbb{R}^M$ at the sensor locations $\{x_m\}_{m=1}^M$. The continuous update field over the spatial domain $\Omega$ is then reconstructed via projection onto the Riesz representers:
\begin{equation}
\eta_{N,M}(x) = \sum_{m=1}^M \eta_{\theta}(x_m) q_m(x)
\end{equation}
where $q_m \in \mathcal{U}$ are the Riesz representers satisfying $(q_m, u) = l_m(u)$ for all $u \in \mathcal{U}$. Each component of $\eta_{\theta}(v)(x_m)$ is associated to a Riesz representer $q_{m}$. Each Riesz representer is associated to a single sensor. For the seek of simplicity, $\eta_{\theta}(v)(x_m)$ will be noted $\eta_{\theta}$.

\subsection{Methodology}

The variational problem reads: find $\left(z_{N, M}, \eta_{\theta}\right) \in \mathcal{Z}_N \times \mathcal{U}_M$ such that
\begin{equation}
\begin{aligned}
\left(\eta_{\theta}, q\right)+\left(z_{N, M}, q\right) &= \left(u^{*}, q\right), && \forall q \in \mathcal{U}_M, \\
\left(\eta_{\theta}, p\right) &= 0, && \forall p \in \mathcal{Z}_N,
\end{aligned}
\label{eq:operator_pbdw_variational}
\end{equation}

In algebraic form, the Operator-PBDW statement reads: find $\left(\mathbf{z}^{\star}, \boldsymbol{\eta}_{\theta}^{\star}\right) \in \mathbb{R}^N \times \mathbb{R}^M$ such that

\begin{equation}
\left[\begin{array}{cc}
\xi M \mathbb{I}_M+\mathbb{A}   &  
 \quad \mathbb{B} \\
\mathbb{B}^T & 0
\end{array}\right]\left[\begin{array}{l}
\boldsymbol{\eta}_{\theta}^{\star} \\
\mathbf{z}^{\star}
\end{array}\right]=\left[\begin{array}{l}
\boldsymbol{y} \\
\mathbf{0}
\end{array}\right]
\label{eq:equation_algebre2}
\end{equation}

\vspace{0.5cm}

In Section 2, it was shown that this problem can be decomposed into an $N$-dimensional constrained optimization problem to find $z_{N,M}$ followed by an $M$-dimensional linear problem to find $\eta_{N,M}$. In this hybrid strategy, the background estimate $z_{N,M}$ is still obtained as the solution of $N$-dimensional constrained optimization problem:
\begin{equation}
\boldsymbol{z}^{\star} = \arg\min_{\boldsymbol{z} \in \mathbb{R}^N} \| \boldsymbol{y} - \mathbb{B}\boldsymbol{z} \|_{\mathbb{W}_{\xi}}^2
\label{eq:background_optimization}
\end{equation}
where $\mathbb{W}_{\xi} \in \mathbb{R}^{M \times M}$ is a positive definite weighting matrix incorporating the regularization parameter $\xi$.

The update term $\eta_{\theta}(v)(x_m)$ is approximated by a neural operator subject to the orthogonality constraint:
\begin{equation}
\mathbb{B}^T \boldsymbol{\eta}_{\theta}^{\star} = \mathbf{0}
\label{eq:orthogonality_constraint}
\end{equation}
This constraint ensures that the neural operator correction remains orthogonal to the background subspace $\mathcal{Z}_N$. The hybrid AI solution combining PBDW with DeepONets is then expressed as:
\begin{equation}
u_{N, M}^* = \sum_{n=1}^N z_n^* \zeta_n + \sum_{m=1}^M \eta_{\theta}(v)(x_m) q_m
\label{eq:hybrid_solution}
\end{equation}
where $\{\zeta_n\}_{n=1}^N$ are the basis functions spanning $\mathcal{Z}_N$, $\{q_m\}_{m=1}^M$ are the Riesz representers, and $\boldsymbol{z}^*$ denotes the optimal background coefficients.  Thus, it is necessary to enforce the orthogonality of the neural operator with the background space. In the following, we present two approaches to enforce orthogonality.

\subsection{Physics-informed DeepONets for bias operator learning - Weak orthogonality}

In this section, we introduce the optimality condition, specifically the orthogonality of the update term with respect to the background space, as a weak constraint within the neural operator framework. We penalize orthogonality violations through a regularization term in the loss function during training.

Recall that a vector is orthogonal to a subspace if and only if it belongs to the orthogonal complement of that subspace. The most natural approach to enforce orthogonality is through inner products. We evaluate the inner product between the DeepONet output $\eta_{\theta}$ and all basis elements $\{\zeta_n\}_{n=1}^N$ spanning the background space $\mathcal{Z}_N$:

\begin{equation}
\mathcal{R}_{\text{orth}}(\eta_{\theta}) = \sum_{n=1}^N \left|(\eta_{\theta}, \zeta_n)\right|^2
\label{eq:orthogonality_regularization}
\end{equation}

Since the background space $\mathcal{Z}_N$ is finite-dimensional with typically small dimension $N$, this computation remains tractable. The orthogonality regularization term $\mathcal{R}_{\text{orth}}$ is then incorporated into the total loss function as a physics-informed constraint:

\begin{equation}
\mathcal{L}_{\text{total}} = \mathcal{L}_{\text{data}} + \lambda_{\text{orth}} \mathcal{R}_{\text{orth}}(\eta_{\theta})
\label{eq:total_loss}
\end{equation}
where $\mathcal{L}_{\text{data}}$ represents the data fidelity term, and $\lambda_{\text{orth}} > 0$ is a hyperparameter controlling the strength of the orthogonality constraint.

\begin{figure}[H]
    \centering
    \includegraphics[width=0.85\linewidth]{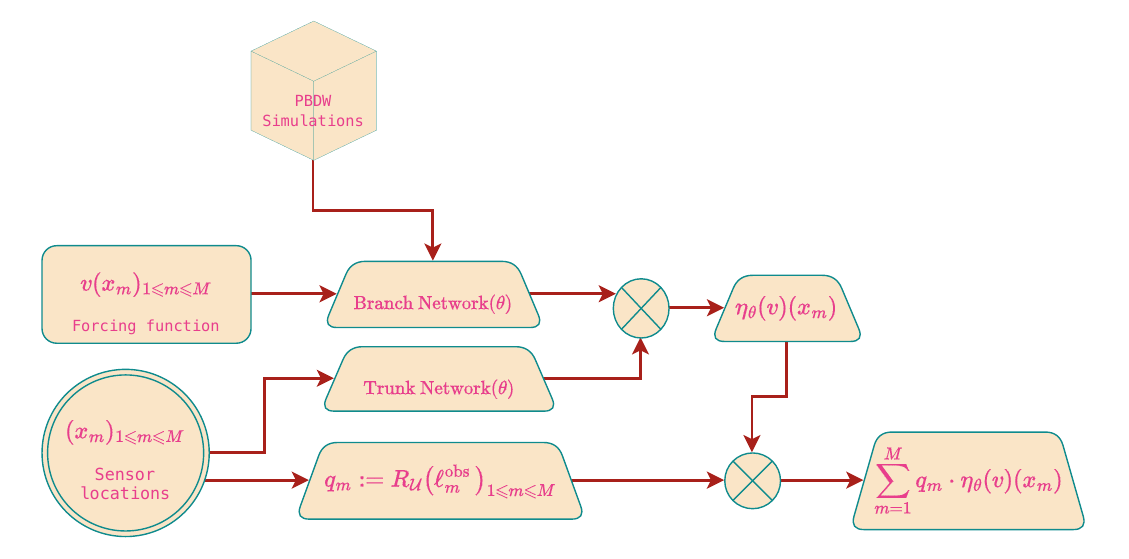}
    \caption{DeepONet with weak constraint.}
    \label{fig:week}
\end{figure}

\begin{table}[h]
\caption{Input and output features}\label{tab2}
\begin{tabular*}{\textwidth}{@{\extracolsep\fill}lcccccc}
\toprule%
& \multicolumn{3}{@{}c@{}}{Branch Network\footnotemark[1]} & \multicolumn{3}{@{}c@{}}{Trunk Network\footnotemark[2]} \\\cmidrule{2-4}\cmidrule{5-7}%
  & Parameters  & & &  Parameters   \\
\midrule
Input  & $v$ &&& $x,y$\\
Output  & $\mathcal{G}(v)$ &&& $\mathcal{T}(v)$ &&\\
\botrule
\end{tabular*}

\footnotetext[1]{ The Branch is implemented as a feed-forward neural network (FNN).}
\footnotetext[2]{ The trunk, implemented as a feed-forward neural network (FNN), takes spatial coordinates $(x,y)$ as input, enabling DeepONet evaluation at arbitrary locations $(x,y)$.}
\end{table}

Let $\theta$ represent all trainable parameters of the neural network and $y$ denote the spatial coordinates at which the training and testing data are evaluated. The optimization problem is formulated as:
\begin{equation}
    \theta^* = \arg \min_{\theta} \mathcal{L}(\theta)
    \label{eq:optimization_problem}
\end{equation}
where the objective function $\mathcal{L}(\theta)$ is defined as:
\begin{equation}
    \mathcal{L}(\theta) = \frac{\omega_1}{N_{\eta}} \sum_{i=1}^{N_{\eta}} \left[ \eta_i(y; \theta) - \eta^*(y) \right]^2 + \frac{\omega_2}{N_{\eta}} \sum_{i=1}^{N_{\eta}} \sum_{j=1}^{N_{\text{modes}}} \left| \left( \eta_i(y; \theta), B_j \right) \right|^2
    \label{eq:loss_function}
\end{equation}

Here, $B_j$ represents the $j$-th column of the coupling matrix $\mathbb{B} \in \mathbb{R}^{M \times N}$ with entries $\mathbb{B}_{mn} = (\zeta_n, q_m)$, where $\zeta_n$ is the $n$-th background mode and $q_m$ is the $m$-th Riesz representer. The model prediction is denoted by $\eta(y; \theta)$, while $\eta^*(y)$ represents the target ground truth. The loss function comprises two weighted terms:
\begin{itemize}
    \item The \textbf{data fidelity term} (weighted by $\omega_1$), which measures the mean squared error between predictions and ground truth;
    \item The \textbf{orthogonality penalty term} (weighted by $\omega_2$), which quantifies violations of orthogonality between the predicted update and the background basis functions $\{B_j\}$.
\end{itemize}

With $\mathcal{G}(v)(x_m)$ being the neural operator evaluation at sensor location $x_m$ as presented in Figure~\ref{fig:week}, the reconstructed continuous scalar field is expressed as:
\begin{equation}
    u_{N, M}^* = \sum_{n=1}^N z_n^* \zeta_n + \sum_{m=1}^M \mathcal{G}(v)(x_m) q_m
    \label{eq:reconstructed_field}
\end{equation}

A key advantage of this approach is that spatial coordinates are parameterized as inputs to the neural operator. Consequently, while the network is trained at fixed sensor locations, it can generate predictions at any point within the spatial domain $\Omega$. It is important to note that the orthogonality constraint is enforced as a \textit{weak constraint} through the penalty term, meaning orthogonality is only approximately satisfied during training and may not hold exactly during inference. This limitation motivates the developments presented in the next section.

\subsection{Physics-constrained DeepONets for bias operator learning - Strong orthogonality}

\begin{figure}[H]
    \centering
    \includegraphics[width=0.85\linewidth]{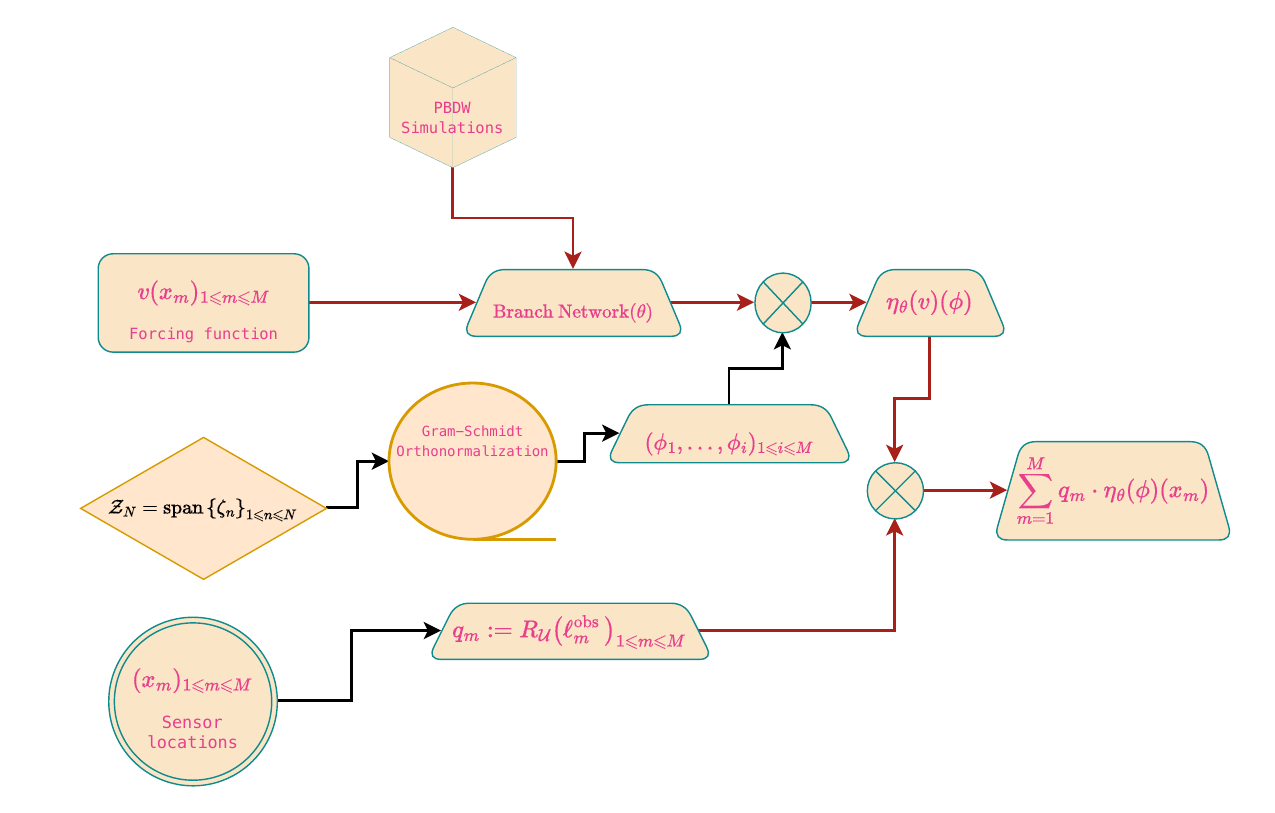}
    \caption{DeepONet with strong constraint.}
    \label{fig:strong}
\end{figure}

In the previous section, weak orthogonality was introduced through penalization in the training stage. However, this does not provide a guarantee of orthogonality outside of the training dataset. Here we propose a method to guarantee the orthogonality constraint in both the training and inference stages (see Figure~\ref{fig:strong}). 

The goal is to capture what $\mathcal{Z}_N = \text{span}\{\zeta_n\}_{n=1}^N$ is missing from the experimental observations. We want the neural operator to lie in $\mathcal{Z}_N^\perp$. Therefore, we modify the trunk network outputs $\{\phi_m\}_{m=1}^M$ to ensure they are orthogonal to all background basis vectors $\{\zeta_n\}_{n=1}^N$.

\subsubsection{Construction via modified Gram-Schmidt process}

Let $\{\phi_m\}_{m=1}^M$ denote the trunk network outputs. We orthogonalize each $\phi_m$ against all background modes $\{\zeta_n\}_{n=1}^N$ by projecting out any components that lie in $\mathcal{Z}_N$. 

Define the projection operator $\Pi$ of a vector $\phi_m$ onto a background basis vector $\zeta_n$ as:
\begin{equation}
\Pi_{\zeta_n} \phi_m = \frac{\langle \phi_m, \zeta_n \rangle}{\langle \zeta_n, \zeta_n \rangle} \zeta_n
\end{equation}

For each trunk output $\phi_m$ (where $m = 1, 2, \ldots, M$), we subtract the projections onto all background basis vectors $\{\zeta_n\}_{n=1}^N$. The orthogonalized trunk basis is computed as:
\begin{equation}
    \phi_m \leftarrow \phi_m - \sum_{n=1}^{N} \Pi_{\zeta_n} \phi_m 
\end{equation}

This operation ensures that each $\phi_m$ is orthogonal to every $\zeta_n$:
\begin{equation}
\langle \phi_m, \zeta_n \rangle = \left\langle \phi_m - \sum_{k=1}^{N} \Pi_{\zeta_k} \phi_m, \zeta_n \right\rangle = \langle \phi_m, \zeta_n \rangle - \langle \phi_m, \zeta_n \rangle = 0
\end{equation}
for all $m \in \{1, \ldots, M\}$ and $n \in \{1, \ldots, N\}$.

By construction, the trunk outputs $\{\phi_m\}_{m=1}^M$ after orthogonalization lie in the orthogonal complement $\mathcal{Z}_N^\perp$ of $\mathcal{Z}_N$ in $\mathcal{U}$.\\

The discrete neural operator output $\eta_{\theta}(v)(y)$ is expressed as a linear combination of the orthogonalized trunk basis:
\begin{equation}
\eta_{\theta}(v)(y) = \sum_{m=1}^{M} \eta_{\theta_m}(v) \cdot \phi_m(y)
\end{equation}
where $\eta_{\theta_m}(v)$ are the coefficients produced by the branch network for input $v$, and $y$ denotes the spatial coordinate.
Since each $\phi_m$ is orthogonal to every vector in $\{\zeta_n\}_{n=1}^N$ by construction, the inner product between $\eta_{\theta}(v)(y)$ and any background basis vector $\zeta_n$ is:
\begin{align*}
    \langle \eta_{\theta}(v), \zeta_n \rangle &= \left\langle \sum_{m=1}^{M} \eta_{\theta_m}(v) \cdot \phi_m, \zeta_n \right\rangle \\
    &= \sum_{m=1}^{M} \eta_{\theta_m}(v) \langle \phi_m, \zeta_n \rangle \\
    &= \sum_{m=1}^{M} \eta_{\theta_m}(v) \cdot 0 \\
    &= 0
\end{align*}

Thus, $\eta_{\theta}(v)$ is orthogonal to every vector in $\{\zeta_n\}_{n=1}^N$. Consequently, the neural operator output lies in $\mathcal{Z}_N^\perp$ by construction, and the optimality condition $\mathbb{B}^T \boldsymbol{\eta}_{\theta}^{\star} = 0$ is enforced exactly—not as a soft penalty, but as a structural constraint embedded in the architecture itself.

\subsubsection{Practical implementation}

In practice, this orthogonalization is performed as a layer operation after the trunk network evaluation:
\begin{enumerate}
    \item The trunk network evaluates $\{\phi_m(y)\}_{m=1}^M$ at spatial locations $y$
    \item For each $\phi_m$, compute projections onto all background modes $\{\zeta_n\}_{n=1}^N$
    \item Subtract these projections to obtain the orthogonalized $\phi_m(y)$
    \item Use the orthogonalized $\{\phi_m(y)\}_{m=1}^M$ in the DeepONet combination with branch outputs
\end{enumerate}

This approach ensures exact orthogonality during both training and inference, removing the need for regularization and the tuning of its hyperparameter.

\begin{table}[h]
\caption{Input and output features}\label{tab2}
\begin{tabular*}{\textwidth}{@{\extracolsep\fill}lcccccc}
\toprule%
& \multicolumn{3}{@{}c@{}}{Branch Network\footnotemark[1]} & \multicolumn{3}{@{}c@{}}{Trunk Network\footnotemark[2]} \\\cmidrule{2-4}\cmidrule{5-7}%
  & Parameters  & & &  Parameters   \\
\midrule
Input  & $v$ &&& $\phi_m$\\
Output  & $\mathcal{G}(v)$ &&& $\phi_m$ &&\\
\botrule
\end{tabular*}
\footnotetext[1]{ the Branch is implemented as a feed-forward neural network (FNN).}
\footnotetext[2]{ the trunk network is replaced by a set of pre-computed basis functions $\{\phi_m\}_{m=1}^M$. Since spatial coordinates $x,y$ are no longer inputs, the DeepONet can only be evaluated at the $M$ pre-defined locations where these basis functions are computed.}
\end{table}

Similarly, the optimization problem to solve is given by:

\begin{equation}
    \theta^* = \arg \min_{\theta} L(\theta),
\end{equation}

with the loss function \( L(\theta) \) defined by: 
\begin{equation}
L(\theta) =\frac{1}{N_{\eta}} \sum_{i=1}^{N_{\eta}} \sum_{j=1}^{N_{POD}}    \left[  \hat{\eta}_i\left(y ; \theta\right) \cdot \phi_{j}-\eta^*\left(y\right)\right]^2 
\end{equation}
Here, $y$ only represents the parametric discretization of the training and testing data. Because the trunk is fixed (replaced by a set of orthonormal basis), both the training and inference are evaluated at fixed locations. This requires the sensors to be optimally selected, which is discussed in the next section.\\

\subsection{Optimal selection of sensors}

To recall the context of this framework, we have $M\geq N$, and in this paper we will deal with $M>>N$. Selecting sensor locations in a smart way minimizes the computational cost by choosing a small value of $M$. In order to best represent the unanticipated physics while using the smallest amount of sensors, the Greedy stability maximization described in Algorithm 1 is the optimal procedure. The idea is to maximize the inf-sup constant $\beta_{N, M}=\inf _{z \in \mathcal{Z}_N} \sup _{q \in \mathcal{U}_M} \frac{(z, q)}{\|z\|\|q\|}$ in a greedy manner while minimizing the approximation error.\\



\begin{algorithm}[htbp]
\caption{SGREEDY ${}_M$ Stability-Maximization Algorithm}
\label{alg:SGREEDY}
\begin{algorithmic}[1]
\State \textbf{Input:} $\left\{\mathcal{Z}_N\right\}_{N=1}^{N_{\max}}$ (background approximation spaces)
\State \textbf{Output:} $\left\{\mathcal{U}_M\right\}_{M=1}^{M_{\max}}$ (experimentally observable update spaces)
\For{$M = 1, \ldots, M_{\max}$}
    \State Set $N = \min\left\{N_{\max}, M\right\}$
    \State \textbf{Compute the least-stable mode:}
    \If{$M > 1$}
        \State $w_{\text{inf}} = \underset{w \in \mathcal{Z}_N}{\arg \inf} \sup_{v \in \mathcal{U}_{M-1}} \frac{(w, v)}{\|w\| \|v\|}$
    \Else
        \State Set $w_{\text{inf}}$ to a normalized basis for $\mathcal{Z}_{N=M=1}$
    \EndIf
    \State \textbf{Compute the associated supremizer:} $v_{\text{sup}} = \Pi_{\mathcal{U}_{M-1}} w_{\text{inf}}$
    \State \textbf{Identify the least well-approximated point:} 
    \State $\tilde{x} = \underset{x \in \Omega}{\arg \sup} \left| \left( w_{\text{inf}} - v_{\text{sup}} \right)(x) \right|$
    \State Set $\mathcal{U}_M = \operatorname{span}\left\{\mathcal{U}_{M-1}, R_{\mathcal{U}} \operatorname{Gauss}(\cdot; \tilde{x}, r_M)\right\}$ \\
    
\EndFor
\end{algorithmic}
\end{algorithm}

\section{Online computational complexity analysis}

In this section, we analyze the \textbf{online computational complexity} of the proposed \textbf{PBDW-DeepONet} approach compared to the original \textbf{PBDW} formulation. 
It is evident that the PBDW-DeepONet method has a higher \textit{offline} cost due to the neural operator training. 
Therefore, the following analysis focuses exclusively on the \textit{online} computational stage, where the state reconstruction is performed from sensor observations.

The online stage of the classical PBDW approach is formulated as the following saddle-point problem:
\begin{equation}
\label{eq:pbdw_system}
\begin{bmatrix}
\xi M \mathbb{I}_M + \mathbb{A} & \mathbb{B} \\
\mathbb{B}^T & 0
\end{bmatrix}
\begin{bmatrix}
\boldsymbol{\eta}^{\star} \\
\mathbf{z}^{\star}
\end{bmatrix}
=
\begin{bmatrix}
\boldsymbol{y} \\
\mathbf{0}
\end{bmatrix},
\end{equation}\\

where $\mathbb{A} \in \mathbb{R}^{M \times M}$, $\mathbb{B} \in \mathbb{R}^{M \times N}$, and $\boldsymbol{y} \in \mathbb{R}^{M}$ is the observation vector. Solving the linear system~\eqref{eq:pbdw_system} requires the inversion of a dense matrix of size $(M+N) \times (M+N)$, leading to a computational complexity of $O\big((M + N)^3\big)$.\\

In the proposed \textbf{PBDW-DeepONet} formulation, the estimation of $\boldsymbol{\eta}_{\theta}^{\star}$ through the trained DeepONet model is achieved in constant time $O(1)$, since it only involves a forward pass through the network. 
The remaining computations are associated with solving for $\mathbf{z}^{\star}$, whose cost can be decomposed into three main steps.\\

\textbf{Step 1: Matrix inversion.}\\  
The matrix $\mathbb{P} \in \mathbb{R}^{M \times M}$ must be inverted, which requires $O(M^3)$ operations.\\

\textbf{Step 2: Matrix–vector and matrix–matrix multiplications.}\\  
Once $\mathbb{P}^{-1}$ is available, the following operations are performed:
\begin{itemize}
    \item Compute $\mathbb{P}^{-1} \boldsymbol{y}$, costing $O(M^2)$;
    \item Compute $\mathbb{B}^T \mathbb{P}^{-1} \boldsymbol{y}$, costing $O(MN)$, where $\mathbb{B} \in \mathbb{R}^{M \times N}$;
    \item Compute $\mathbb{B}^T \mathbb{P}^{-1} \mathbb{B}$, costing $O(M^2 N)$.
\end{itemize}

\textbf{Step 3: Solving for $\mathbf{z}^{\star}$} \\ 
Finally, the reduced system $\mathbb{B}^T \mathbb{P}^{-1} \mathbb{B} \mathbf{z}^{\star} = \mathbb{B}^T \mathbb{P}^{-1} \boldsymbol{y}$ is solved, which requires $O(N^3)$ operations. Hence, the total online complexity of the PBDW-DeepONet approach is $O(M^3 + M^2 N + N^3)$

The online complexities of both methods are summarized in Table~\ref{tab:complexity_comparison}.

\begin{table}[h!]
\centering
\caption{Online(Inversion) computational complexity of PBDW and PBDW-DeepONet.}
\label{tab:complexity_comparison}
\begin{tabular}{|l|l|}
\hline
\textbf{Method} & \textbf{Online Complexity} \\
\hline
PBDW-DeepONet & $O(M^3 + M^2 N + N^3)$ \\
\hline
Original PBDW & $O((M + N)^3) = O(M^3 + 3M^2 N + 3M N^2 + N^3)$ \\
\hline
\end{tabular}
\end{table}

From an asymptotic standpoint, both approaches exhibit similar scaling behavior, particularly when $M \gg N$ or $M \approx N$. 
However, in practice, the \textbf{PBDW-DeepONet} method requires fewer online operations than the classical \textbf{PBDW} formulation for the state prediction task. 
This efficiency gain arises from the replacement of one component of the saddle-point system by the neural operator’s fast, learned approximation.

\section{Numerical results }

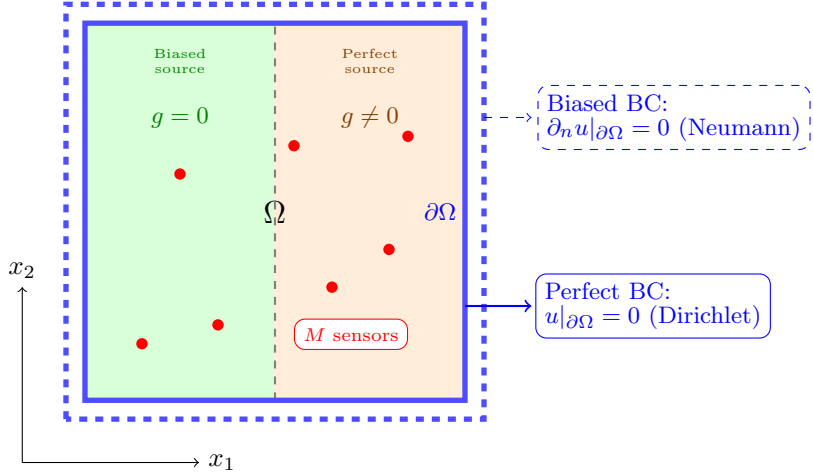
\begin{figure}[h]
\centering
\begin{tikzpicture}[scale=2.5]

\draw[thick, fill=black!30] (0,0) rectangle (2,2);

\fill[green!15] (0,0) rectangle (1,2);
\node[green!50!black, font=\tiny, align=center] at (0.5,1.8) {Biased\\source};

\fill[orange!15] (1,0) rectangle (2,2);
\node[orange!50!black, font=\tiny, align=center] at (1.5,1.8) {Perfect\\source};

\draw[dashed, thick, gray] (1,0) -- (1,2);

\node[font=\large] at (1,1) {$\Omega$};

\draw[blue!70, line width=2pt] (0,0) rectangle (2,2);

\draw[blue!70, line width=2pt, dashed]
    (-0.1,-0.1) rectangle (2.1,2.1);

\node[blue, font=\small] at (1.87,1) {$\partial \Omega$};

\draw[blue, ->, thick] (2.0, 0.5) -- (2.35, 0.5);
\node[blue, font=\scriptsize, align=left, draw, rounded corners, fill=white]
    at (2.99, 0.5) {
    Perfect BC:\\$u|_{\partial\Omega} = 0$ (Dirichlet)
};


\draw[blue, ->, dashed] (2.1, 1.5) -- (2.35, 1.5);
\node[blue, font=\scriptsize, align=left, draw, rounded corners, fill=white, dashed] at (3.1, 1.5) {
    Biased BC:\\$\partial_n u|_{\partial\Omega} = 0$ (Neumann)
};

\draw[->] (-0.33,-0.33) -- (0.6,-0.33) node[right] {$x_1$};
\draw[->] (-0.33,-0.33) -- (-0.33,0.6) node[above] {$x_2$};

\node[green!50!black, font=\scriptsize, align=center] at (0.5, 1.5) {$g = 0$};
\node[orange!50!black, font=\scriptsize, align=center] at (1.5, 1.5) {$g \neq 0$};

\foreach \pos in {(0.3,0.3), (0.7,0.4), (1.3,0.6), (1.6,0.8), (0.5,1.2), (1.1,1.35), (1.7,1.4)} {
    \fill[red] \pos circle (0.03);
}
\node[red, font=\footnotesize, draw, rounded corners, fill=white] at (1.4,0.35) {$M$ sensors};

\end{tikzpicture}

\caption{Schematic representation of the computational domain $\Omega$ with boundary $\partial\Omega$. Two boundary condition scenarios are illustrated: solid lines denote the perfect Dirichlet BC ($u=0$), while dashed lines indicate the misspecified (biased) Neumann BC ($\partial_n u=0$). The color-coded domain partition distinguishes between the biased source formulation (green, $g=0$) and the perfect source formulation (orange, $g\neq 0$). Red circles represent the $M$ measurement sensor locations.}
\label{fig:domainomega}
\end{figure}

We study the Helmholtz equation in a 2D homogeneous medium $\Omega = [0, 1]^2$ with a fixed dissipative phenomenon characterized by the magnitude $\epsilon= 0.01$. The geometry is illustrated in Figure~\ref{fig:domainomega}, where we denote by $\mu$ the wave number. We consider $M$ perfect pointwise observations of the solution field.


The governing equation is the dissipative Helmholtz equation:
\begin{equation}\label{eq:helmholtz}
\begin{cases}
- (1 + \epsilon \mu i) \Delta u(x) - \mu^2 u(x) = \mu q(x) & \forall x \in \Omega, \\
\mathcal{B}(u)(x) = 0 & \forall x \in \partial \Omega
\end{cases}
\end{equation}
where the complex coefficient $(1 + \epsilon \mu i)$ introduces dissipation into the system.

The source term is defined as:
\begin{equation}\label{eq:source}
q\left(x_1, x_2\right) = \sin(\mu x_1) \sin(\mu x_2) + \textcolor{orange}{g(x_1, x_2)}
\end{equation}
where \textcolor{orange}{$g$} represents the bias associated with imperfections in modeling the forcing term. We consider two scenarios:
\begin{equation}\label{eq:bias}
\textcolor{orange}{g\left(x_1, x_2\right)} := 
\begin{cases}
0 & \textcolor{green!50!black}{\text{(biased source model)}} \\ 
-0.5 \sin\left(e^{-3x_1} \sin \left(5\mu x_2\right)\right) & \textcolor{orange}{\text{(perfect source model)}}
\end{cases} 
\end{equation}

The boundary operator $\textcolor{blue}{\mathcal{B}(u)}$ represents two possible boundary conditions:
\begin{equation}\label{eq:bc}
\textcolor{blue}{\mathcal{B}(u)(x)} :=
\begin{cases} 
u(x) & \text{(Dirichlet: perfect BC, shown as \textbf{solid} line)} \\ 
\partial_n u(x) & \text{(Neumann: misspecified BC, shown as \textbf{dashed} line)}
\end{cases} 
\quad \forall x \in \partial \Omega
\end{equation}
These scenarios allow us to study the impact of model errors on the inverse problem.


\subsection{Reference solution}

In this study, we use a finite element solution as a high-fidelity reference, or ground truth $u_{\text {true }}$. The underlying boundary value problem is well-posed, guaranteeing the existence and uniqueness of the solution \( u \).

The problem is formulated in its  weak form:
\[
\text{find } u \in \mathcal{U} \text{ such that } a(u, v) = l(v), \quad \forall v \in \mathcal{U}.
\]
where
\begin{align}
a(u,v) &= \int_{\Omega} \left( -(1+i\varepsilon) \nabla u \cdot \nabla \bar{v} - \mu^2 u \bar{v}\right  ) \mathrm{d}\Omega + \text{boundary terms}, \\
\ell(v) &= \int_{\Omega} \mu q \bar{v} \, \mathrm{d}\Omega + \text{boundary terms}.
\end{align}

To compute a numerical approximation, we discretize the domain \( \Omega \) into a mesh and define \( \mathcal{U}_h \subset \mathcal{U} \) as a finite-dimensional subspace of piecewise polynomial functions.

The approximate solution \( u_h \in \mathcal{U}_h \) is then determined by solving the Galerkin projection of the continuous problem:
\[
\text{find } u_h \in \mathcal{U}_h \text{ such that } a(u_h, v_h) = \ell(v_h) \quad \forall v_h \in \mathcal{U}_h.
\]

The resulting \( u_h \) is our reference solution against which other methods are compared.

\subsection{Background and update spaces}

As in [1] the experimentally observable space is defined as $\mathcal{U}_M:=\operatorname{Span}\left\{q_m\right\}_{1 \leqslant m \leqslant M} \subset \mathcal{U}$, where $q_m:=R_{\mathcal{U}}\left(\ell_m^{\text {obs }}\right)$ is the Riesz representation of $\ell_m^{\text {obs }} \in \mathcal{U}^{\prime}$, with $\mathcal{U}^{\prime}$ the dual of $\mathcal{U}$. The experimental observations of the true state satisfy $\left(u^{\text {true }}, q_m\right)=\ell_m^{\text {obs }}\left(u^{\text {true }}\right)$, for all $m \in\{1, \ldots, M\}$.\\

In this case, the sensors are modeled through a Gaussian function such that: $ \ell_m^{\mathrm{obs}}(v)=\frac{1}{\sqrt{2 \pi r_m^2}} \int_{\mathcal{R}_m} v(x) \exp \left(\frac{-\left(x-x_m^c\right)^2}{2 r_m^2}\right) d x
$
where $x_m^c$ is the center of the Gaussian function that reflects the location of the sensor, and $r_m \ll\left|\mathcal{R}_m\right|^{\frac{1}{d}}$ is the standard deviation of the Gaussian function that reflects the filter width of the sensor.\\

The background space $\mathcal{Z}_N$ is the condensation of the best knowledge of $\mathcal{M}^{\text {bk }}$ into a $N$-dimensional linear space $\mathcal{Z}_N$ through a model reduction process: $\operatorname{Process}_N^{\mathcal{Z}}\left(\mathcal{M}^{\mathrm{bk}}\right) \rightarrow \mathcal{Z}_N$. Here we use the Proper orthogonal decomposition (POD): $\operatorname{ProcesS}_N^{\mathcal{Z}}\left(\mathcal{M}^{\mathrm{bk}}\right) \equiv \operatorname{POD}_N\left(\mathcal{M}^{\mathrm{bk}}\right)$.
We first introduce a training set $\Xi_{\text {train }} \subset \mathcal{D}$ that sufficiently covers the parameter domain $\mathcal{D}$. We then evaluate the best-knowledge solution (defined from the biased source and boundary conditions) at each training point to form the set $\left\{u^{\text {bk }, \mu}\right\}_{\mu \in \Xi_{\text {train }}}$. We finally apply POD to $\left\{u^{\text {bk, } \mu}\right\}_{\mu \in \Xi_{\text {train }}}$ and extract the $N$ most dominant modes as measured in $\|\cdot\|$ to form $\mathcal{Z}_N$.

\begin{figure*}[htp]
    \centering

    \subfloat[$R_{\mathcal{U}}(\ell_1^{\text{obs}})$]{
        \includegraphics[width=0.32\textwidth]{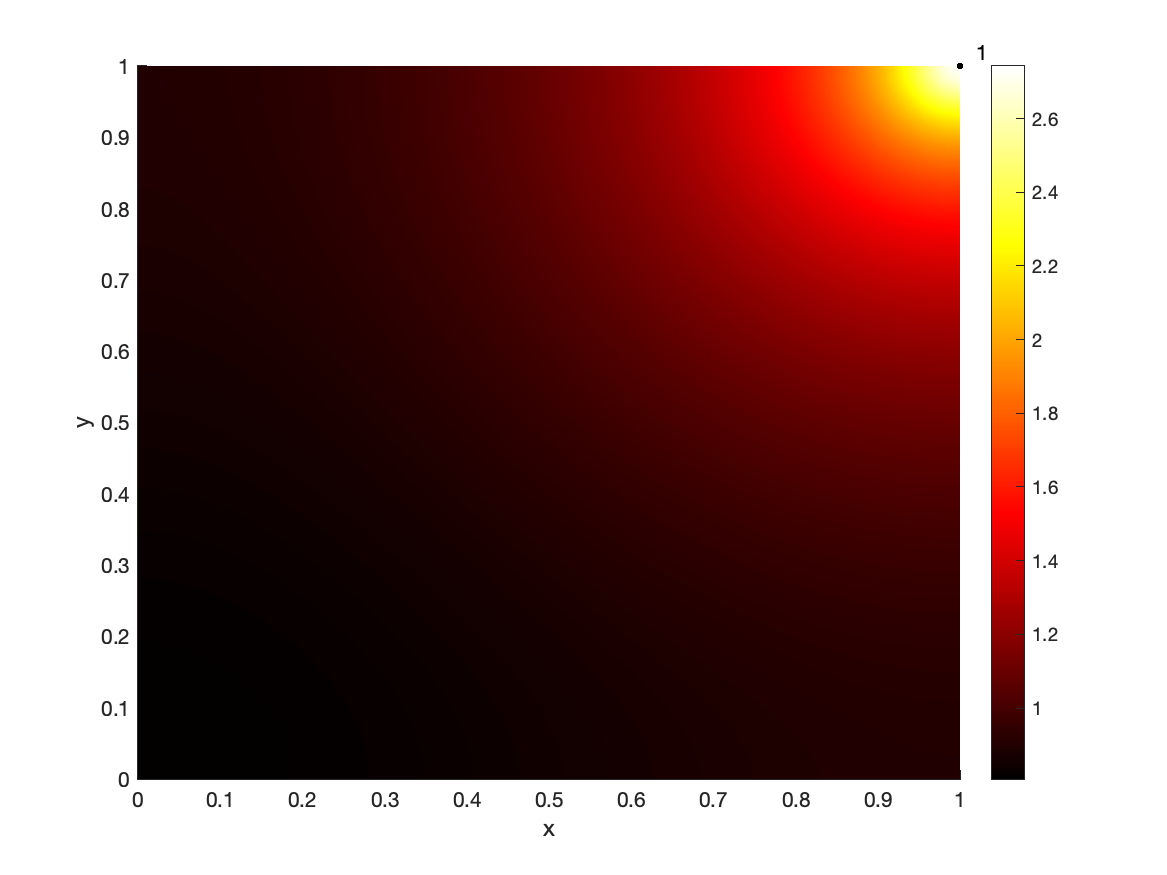}}
    \subfloat[$R_{\mathcal{U}}(\ell_2^{\text{obs}})$]{
        \includegraphics[width=0.32\textwidth]{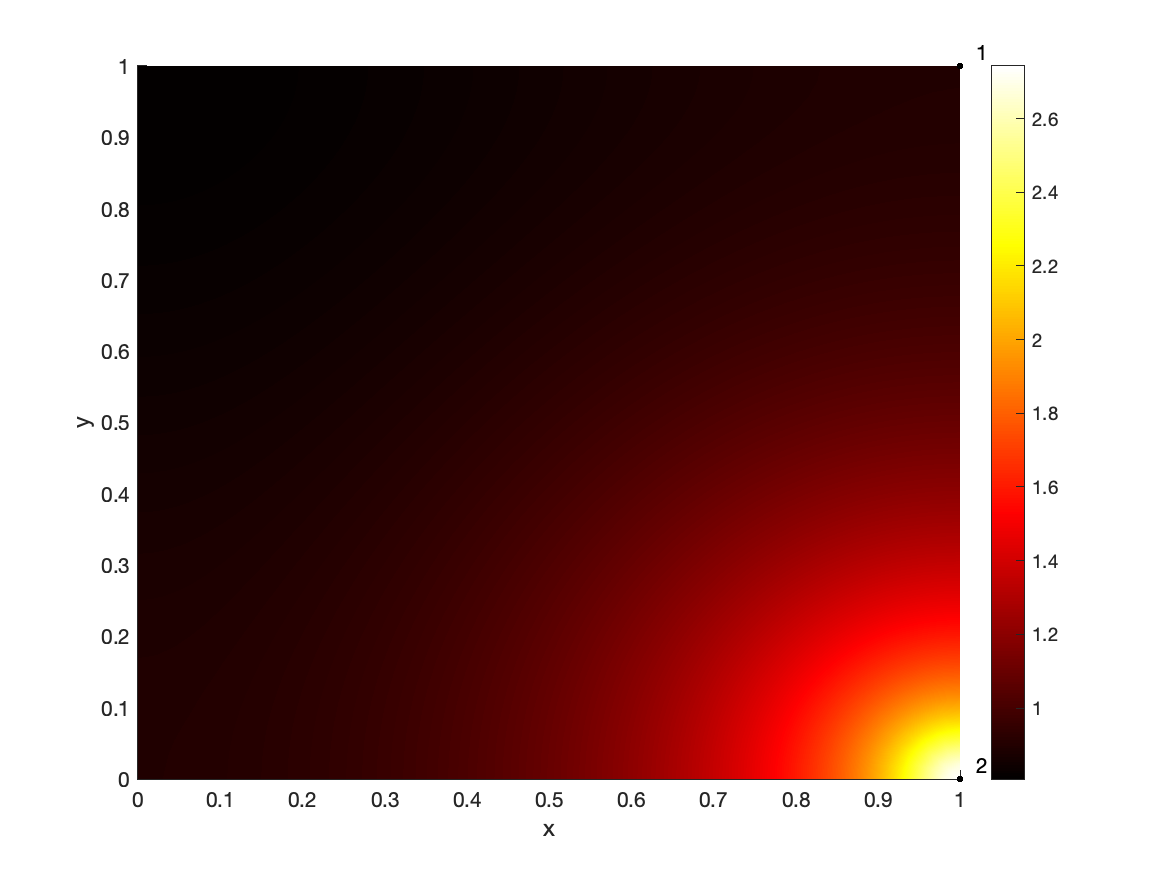}}
    \subfloat[$R_{\mathcal{U}}(\ell_3^{\text{obs}})$]{
        \includegraphics[width=0.32\textwidth]{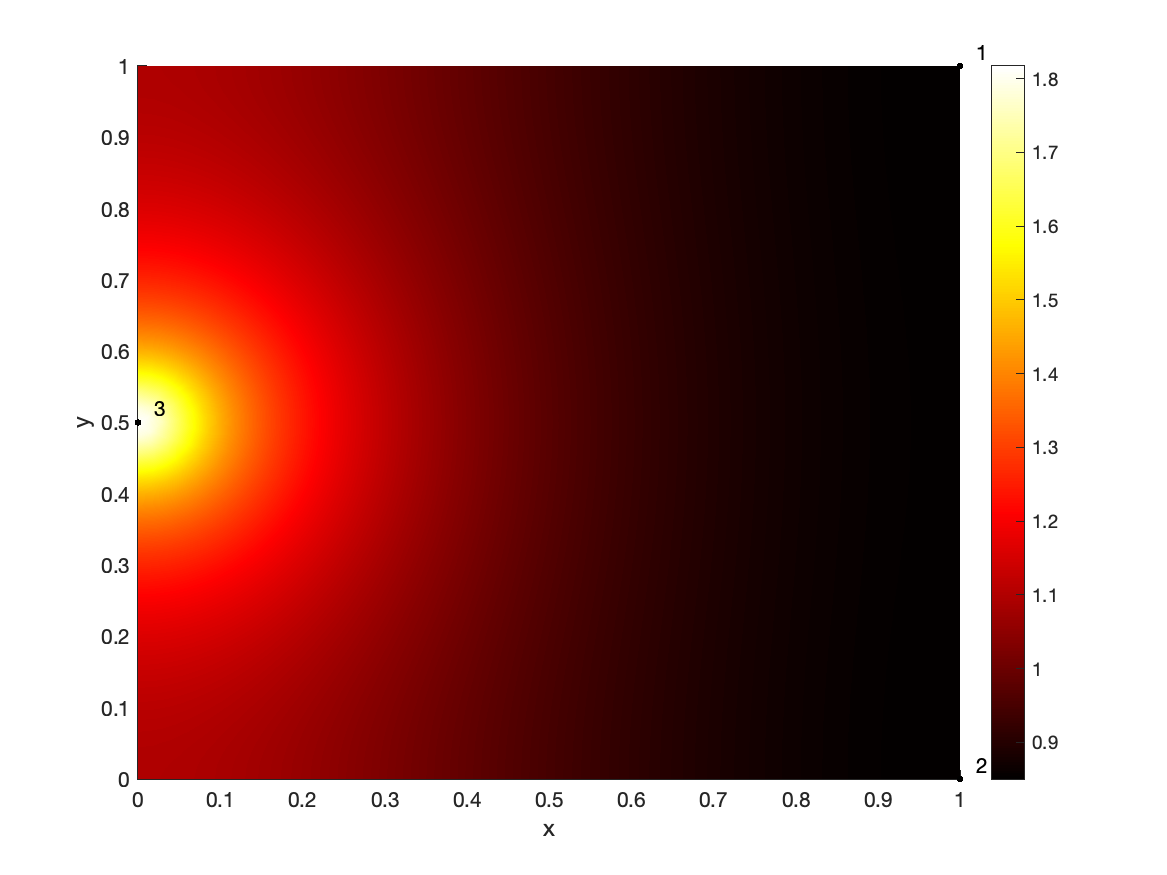}}

    \vspace{0.4cm}

    \subfloat[$R_{\mathcal{U}}(\ell_4^{\text{obs}})$]{
        \includegraphics[width=0.32\textwidth]{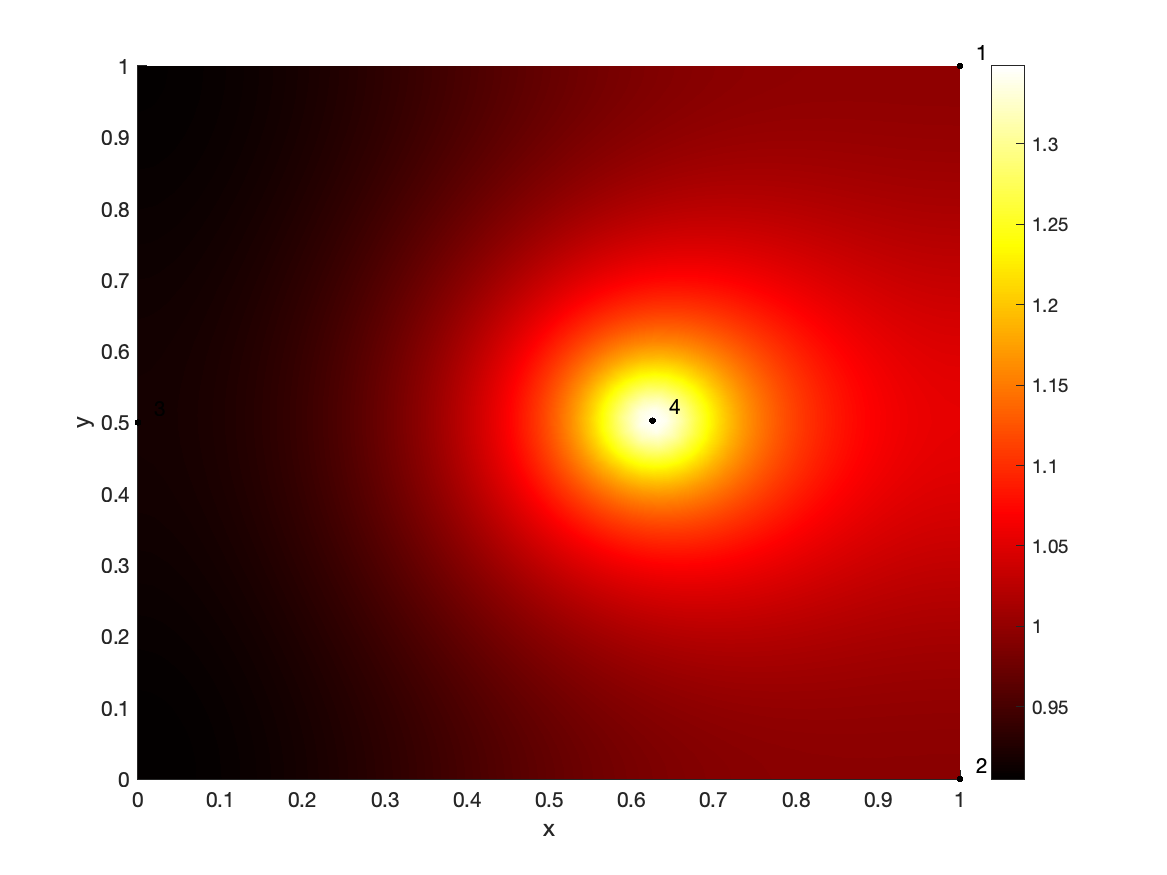}}
    \subfloat[$R_{\mathcal{U}}(\ell_5^{\text{obs}})$]{
        \includegraphics[width=0.32\textwidth]{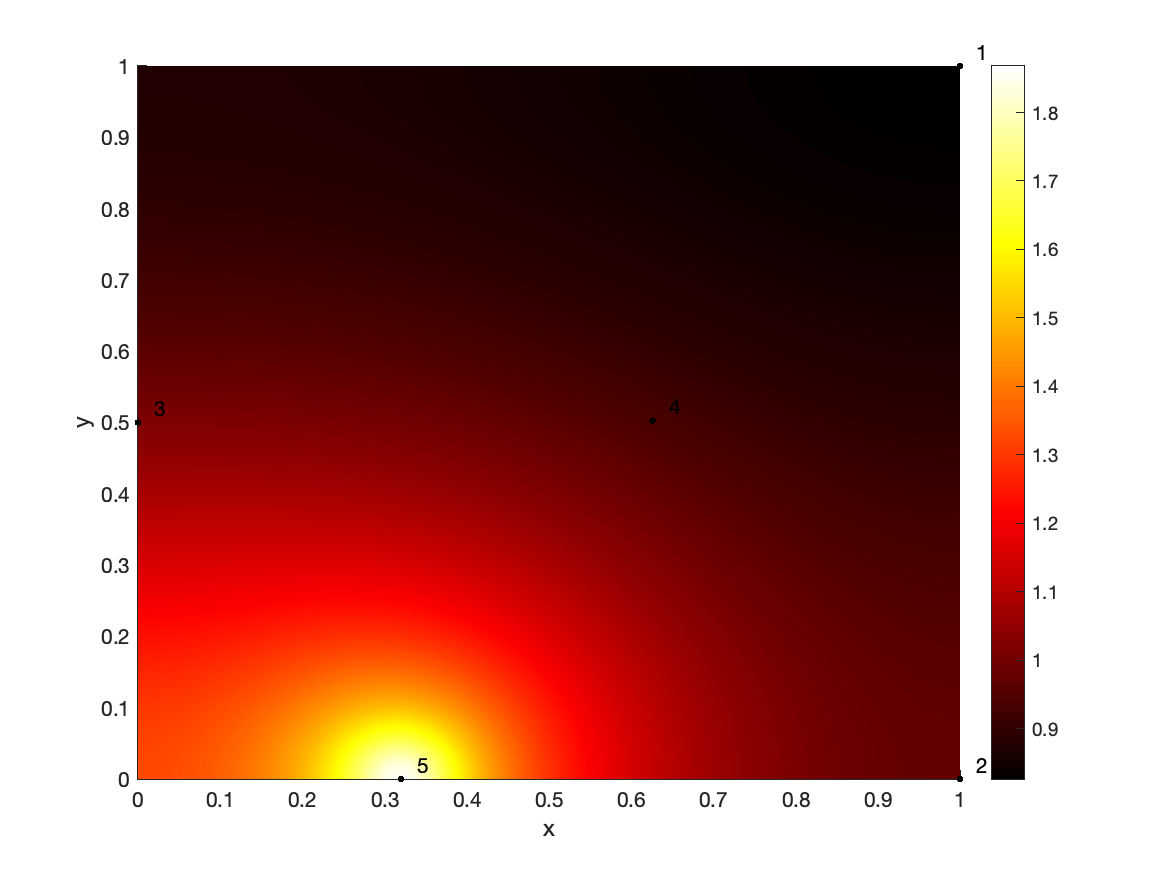}}
    \subfloat[$R_{\mathcal{U}}(\ell_6^{\text{obs}})$]{
        \includegraphics[width=0.32\textwidth]{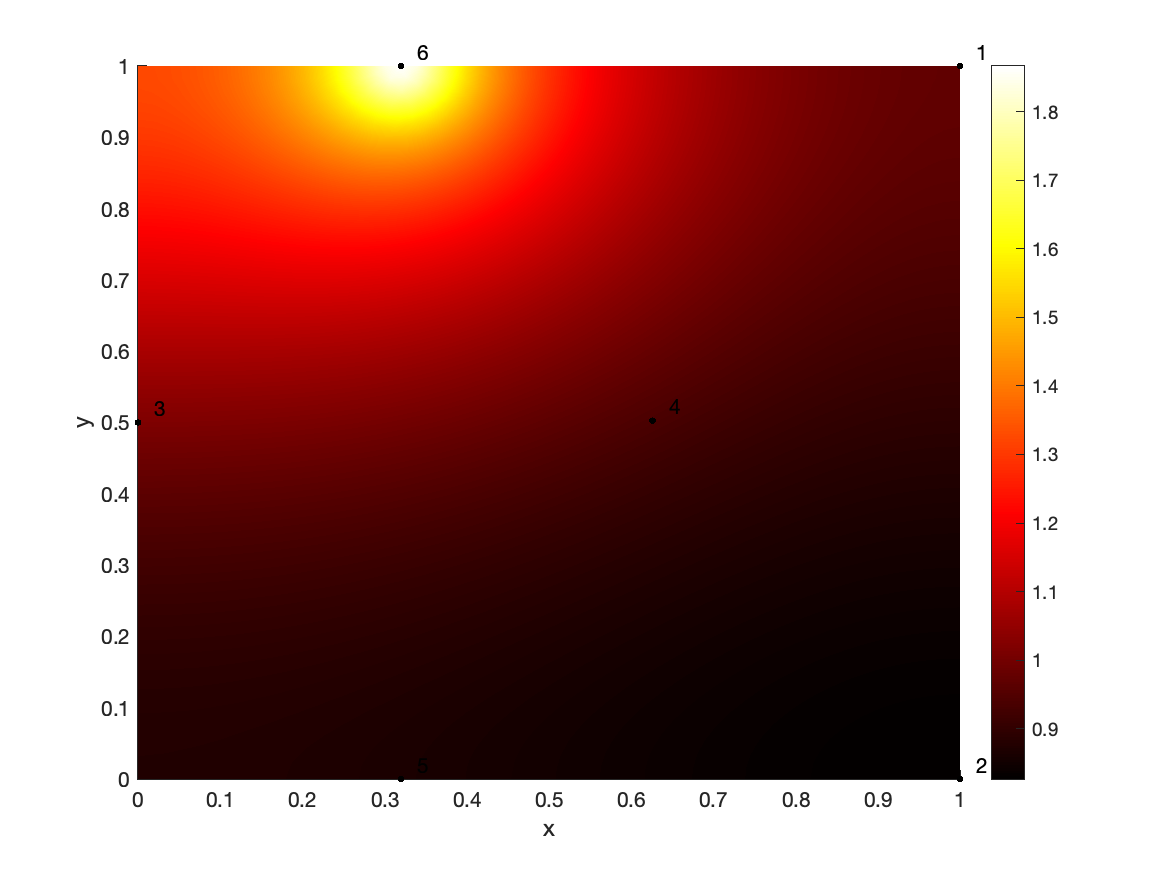}
    }

    \caption{Observation space $\mathcal{U}_M := \operatorname{Span}\left\{q_m\right\}_{1 \leqslant m \leqslant M}$}
    \label{fig:nbRayons}
\end{figure*}

\begin{figure*}[htp]
    \centering

    \subfloat[$\zeta_1$]{
        \includegraphics[width=0.32\textwidth]{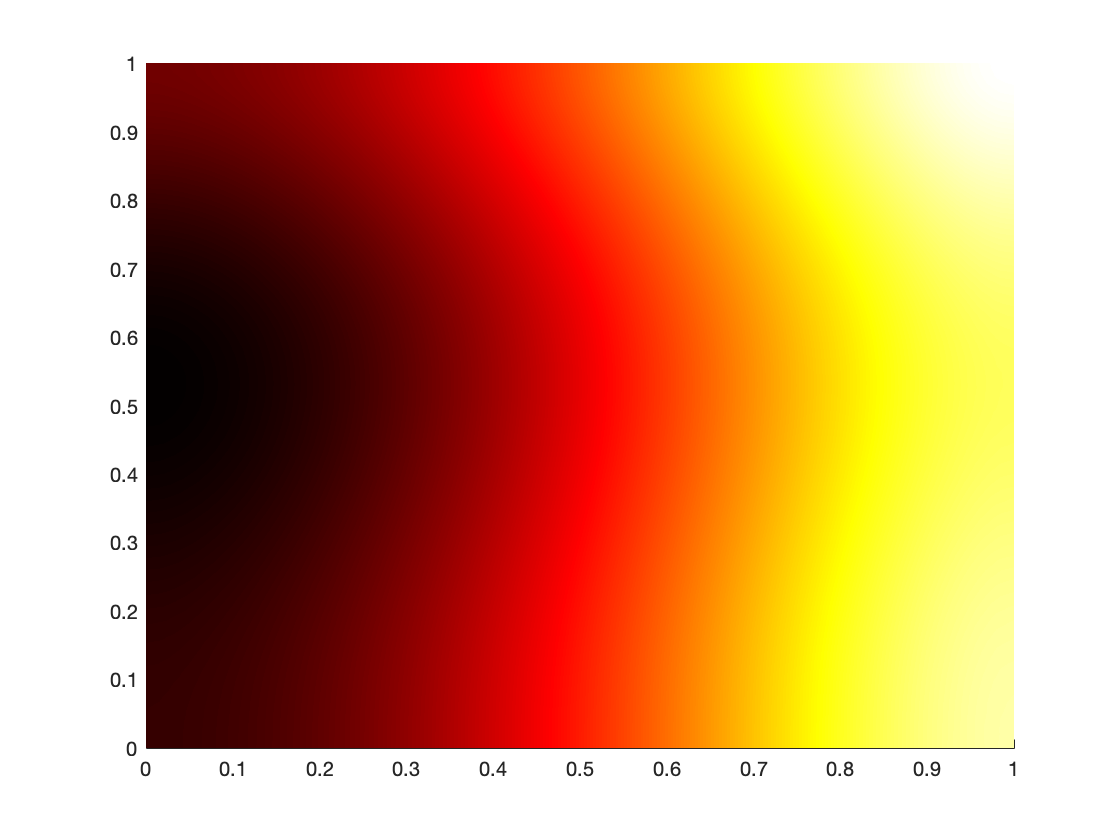}}
    \subfloat[$\zeta_2$]{
        \includegraphics[width=0.32\textwidth]{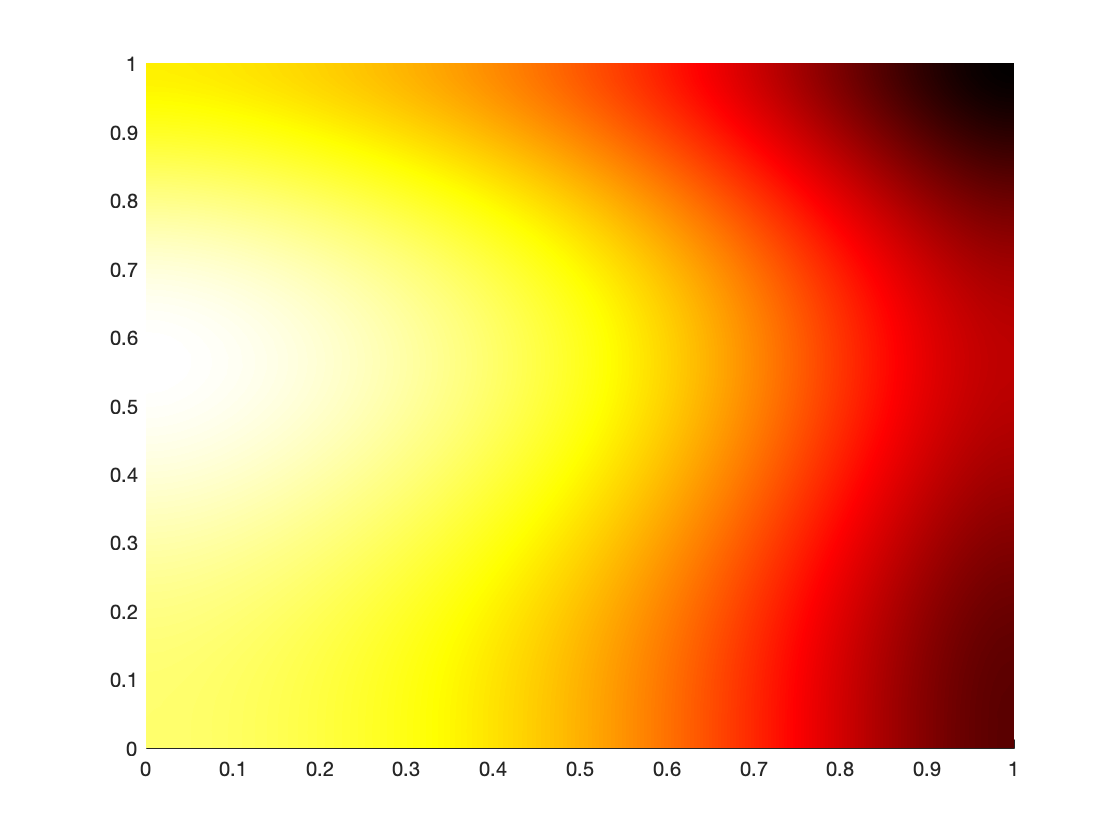}}
    \subfloat[$\zeta_3$]{
        \includegraphics[width=0.32\textwidth]{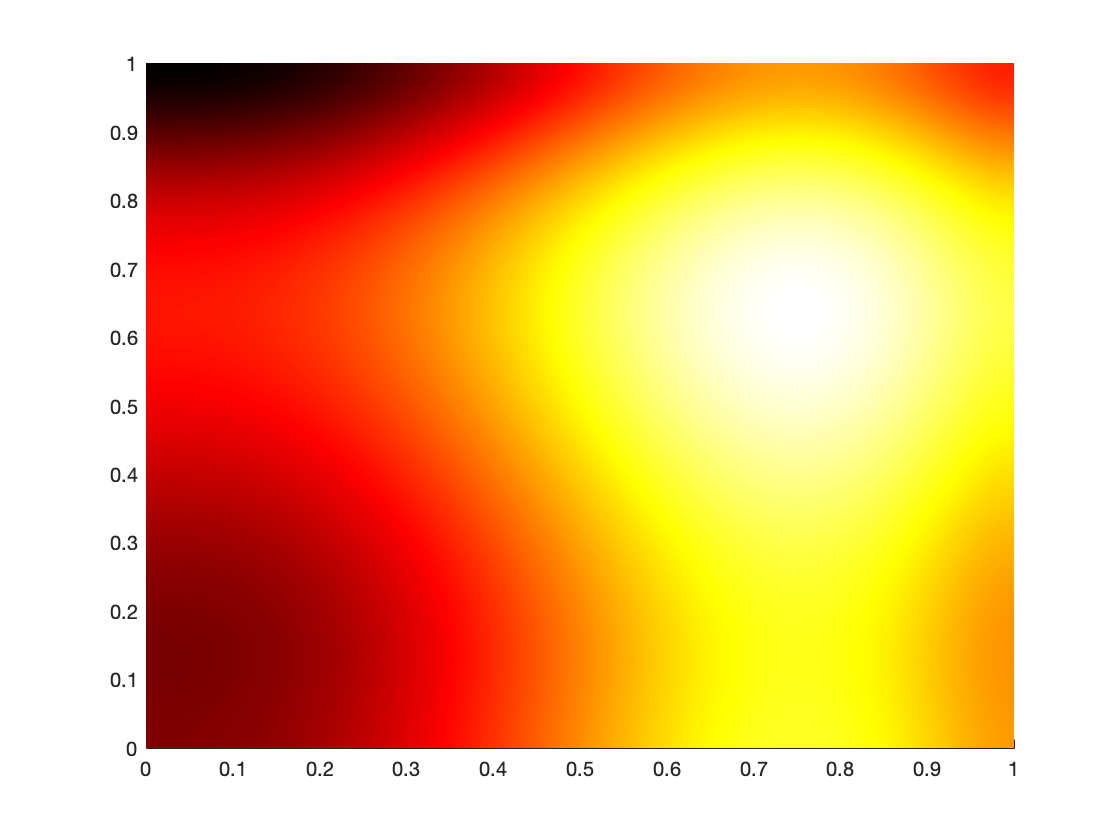}
    }

    \vspace{0.4cm}

    \subfloat[$\zeta_4$]{
        \includegraphics[width=0.32\textwidth]{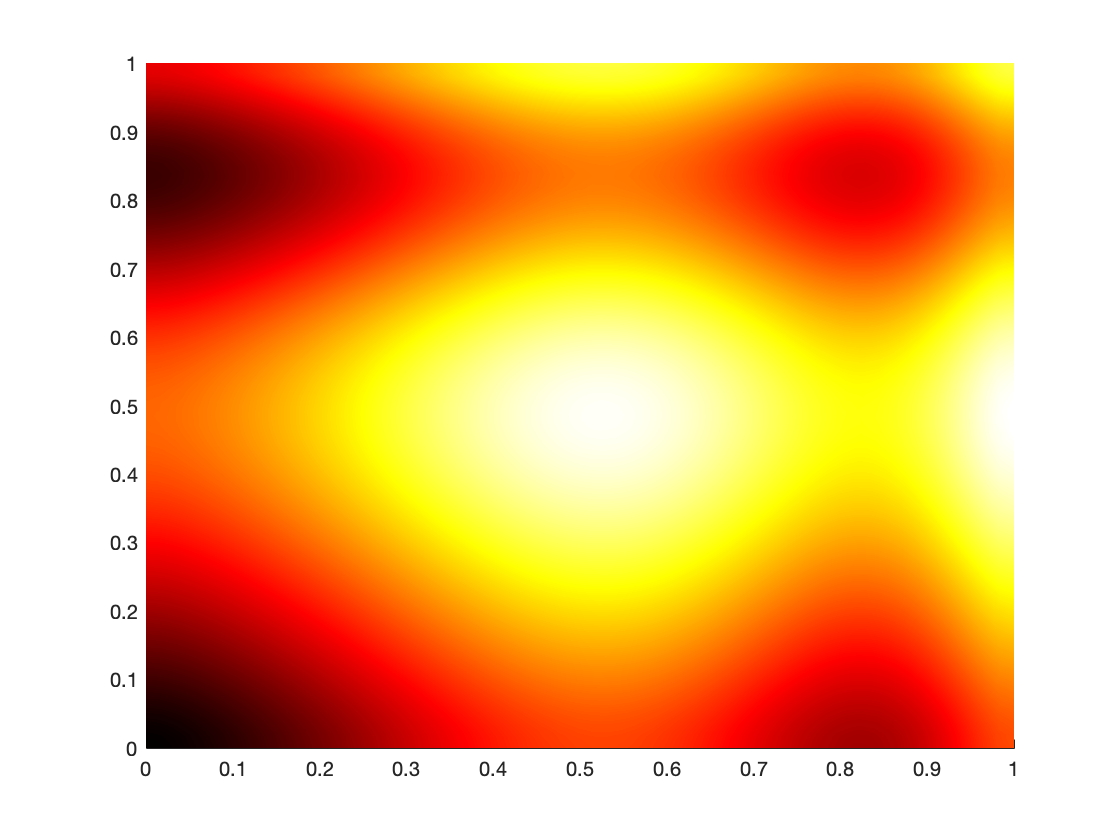}}
    \subfloat[$\zeta_5$]{
        \includegraphics[width=0.32\textwidth]{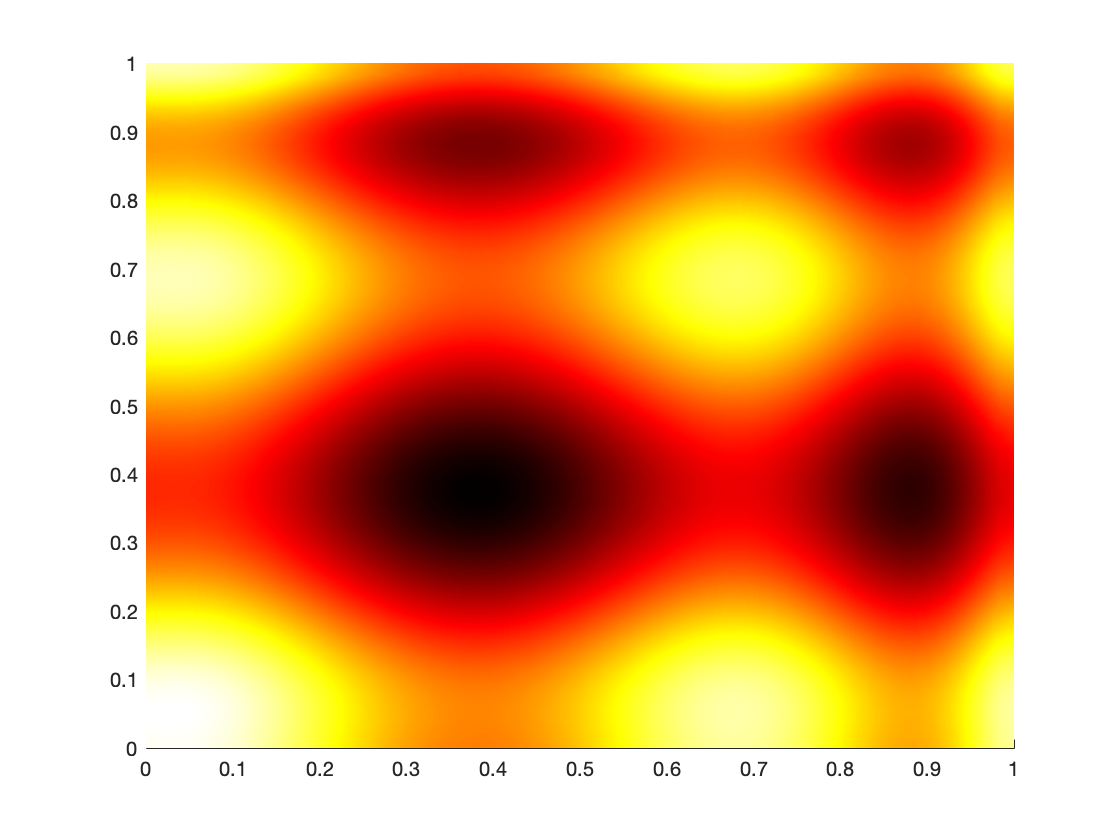}}
    \subfloat[$\zeta_6$]{
        \includegraphics[width=0.32\textwidth]{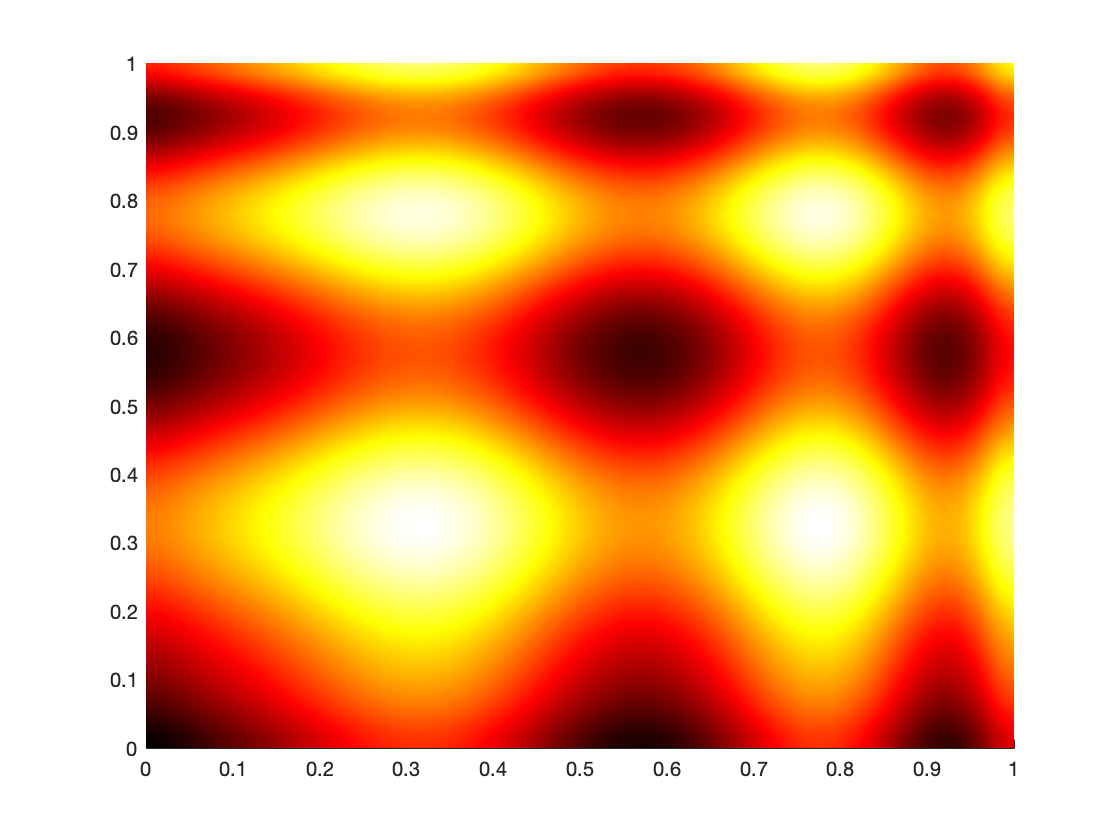}}

    \caption{Background space $\mathcal{Z}_N = \operatorname{Span}\{\zeta_n\}_{1 \le n \le N}$.}
    \label{fig:nbRayons}
\end{figure*}

\subsection{Architecture of the neural network}

In both the weak and strong orthogonality approaches, the branch net is a feedforward neural network with $10$ layers, $M$ neurons peer layer in the hidden layers, $M$ neurons in the output layer, with $M$ the number of sensors. Specifically,
in the weak case, the trunk is parameterized by a feedforward neural network with 4 hidden layers and $M$ neurons per layer. In the strong constraint case, the trunk net is a tensor of size $M*M$, with $M$ orthonormal basis functions of dimension $M$, each selected by the stability maximization algorithm. The Pytorch module was used for all the experiments.\\

In the weak constraint case, 501 pairs of input/output data were used for the learning. The amount of data used to train the DeepONet is highly reduced in the strong constraint case as the trunk being replaced by a set of basis functions allows a faster learning, and the optimal selection of sensors reduces the dimension of the update space thus the number parameters of the neural network to be trained. For the strong constraint only 50 pairs of input/output data were used for this supervised task. The mean squared error (MSE) between the operator prediction and the ground truth was minimized using Adam optimization with a learning rate decay. 

\subsection{Training curves of the neural operators }

\begin{figure}[H]
    \centering
    \includegraphics[width=0.75\linewidth]{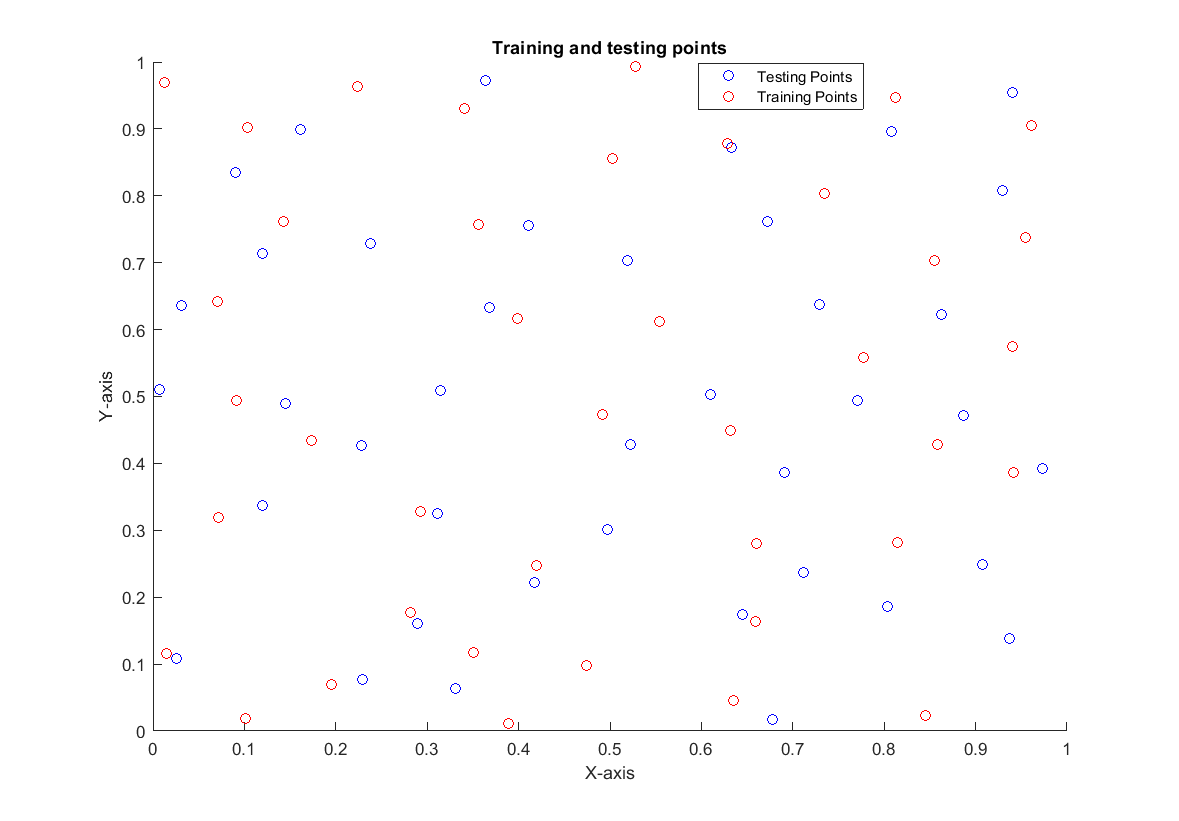}
    \caption{Training and testing points.}
    \label{fig:random_sensors}
\end{figure}

Two neural operator models are trained under different sensor locations strategies.  The first strategy, associated with the \textit{weak-orthogonality} formulation, employs a random selection of sensor locations since this model can predict at any spatial point.  In this approach, the positions of the $M$ sensors are uniformly distributed while maintaining a minimum inter-sensor distance of $1 / \sqrt{3M}$, ensuring adequate spatial dispersion across the entire domain. Specifically, the model is trained on the blue points and tested on the red points (Figure ~\ref{fig:random_sensors}).

The second strategy leverages a sensor-placement procedure guided by the \textit{inf-sup} constant $\beta$ (S-Greedy strategy), aiming to optimize sensor informativeness and numerical stability as detailed in Section 3.4. 

\begin{figure*}[htp]
   \subfloat[Weak orthogonality]{\label{rev}
      \includegraphics[width=.505\textwidth]{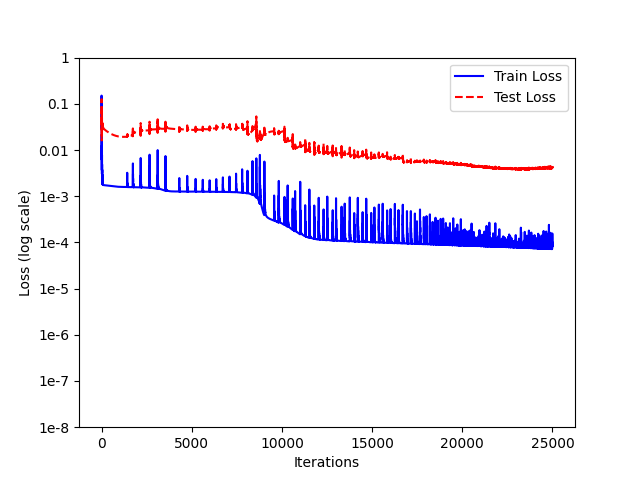}}
~
   \subfloat[Strong orthogonality]{\label{rev_sol}
      \includegraphics[width=.47\textwidth]{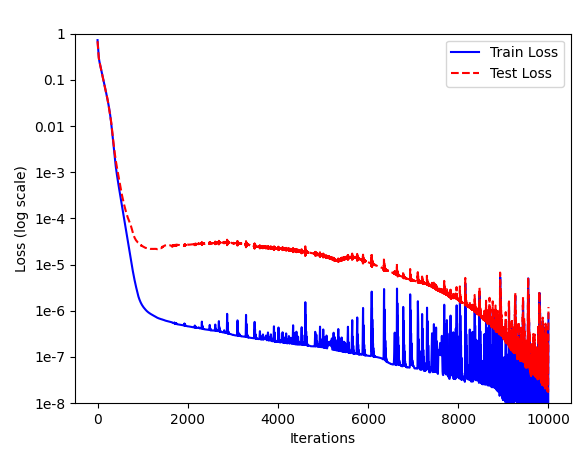}}

   \caption{Evolution of training and testing loss for the weak constraint approach. }\label{2loss}
\end{figure*}

Figure~\ref{2loss} shows that the second approach, which enforces strong orthogonality, demonstrates superior training performance of the neural operator in terms of mean squared error (MSE) compared to the weak-constraint formulation.

\subsection{Quantity of interest}

In order to assess performance of our approach, we will focus on several metrics:\\

\begin{itemize}

    \item The recovery error $e_{\text {exact }}=\left\|u_{\text {true }}-u_{N, M}\right\|_{L_2}^2$ : this error quantifies the estimation by the PBDW.\\

    \item The background estimate error $e_{\text {estim }}=\left\|u_{\text {true }}-z_{N, M}\right\|_{L_2}^2$ : the evaluation of the model's tendency to approach the solution considers both the model error and the approximation error.\\
 
    \item The update norm $\left\|\eta_{N, M}\right\|_{L^2}$ : it characterizes the importance of the correction in the reconstruction, taking into account model deficiencies (approximation error and model bias) and measurement noise. This quantity can be calculated without the need for knowledge of the exact solution.\\
    

    \item The projection error of the reduced basis, $e_{S V D}$. It should be noted that if the estimation error follows the trend of the projection error, this implies that the model bias is negligible. This quantity can be calculated in the offline phase, thus providing us with a priori information.

\end{itemize}

\subsection{Perfect model without regularization}

This is an ideal case: perfect measurement (absence of noise) and a perfect model (absence of bias). This case will be
used to validate the PBDW, in particular to determine the number of reduced modes $N$ in the background space needed to approximate the exact solution.
Increasing the number of modes induces an ill-posedness amplification; with more background modes, there are more directions in the solution space that may be poorly observed by the fixed set of measurements. Some of these additional modes might be nearly invisible to the observation operator, leading to decreased stability.

\begin{figure}[H]
    \centering
    \includegraphics[width=0.95\linewidth]{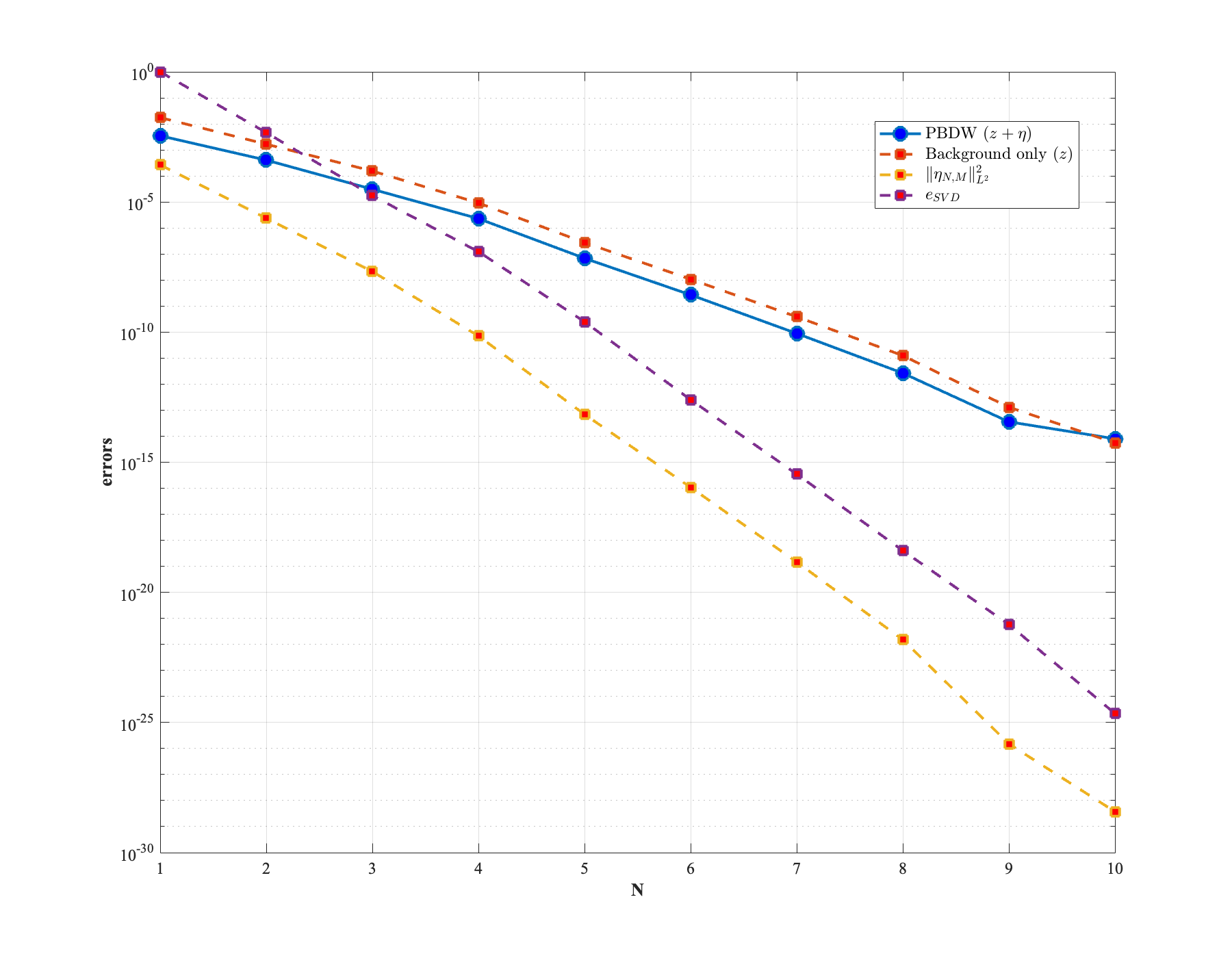}
    \caption{Evolution of errors as a function of the number $N$ of modes.}
    \label{fig:global_errors}
\end{figure}

\begin{figure*}[htp]
    \subfloat[PBDW $z_{2,50}$]{
        \includegraphics[width=0.25\textwidth]{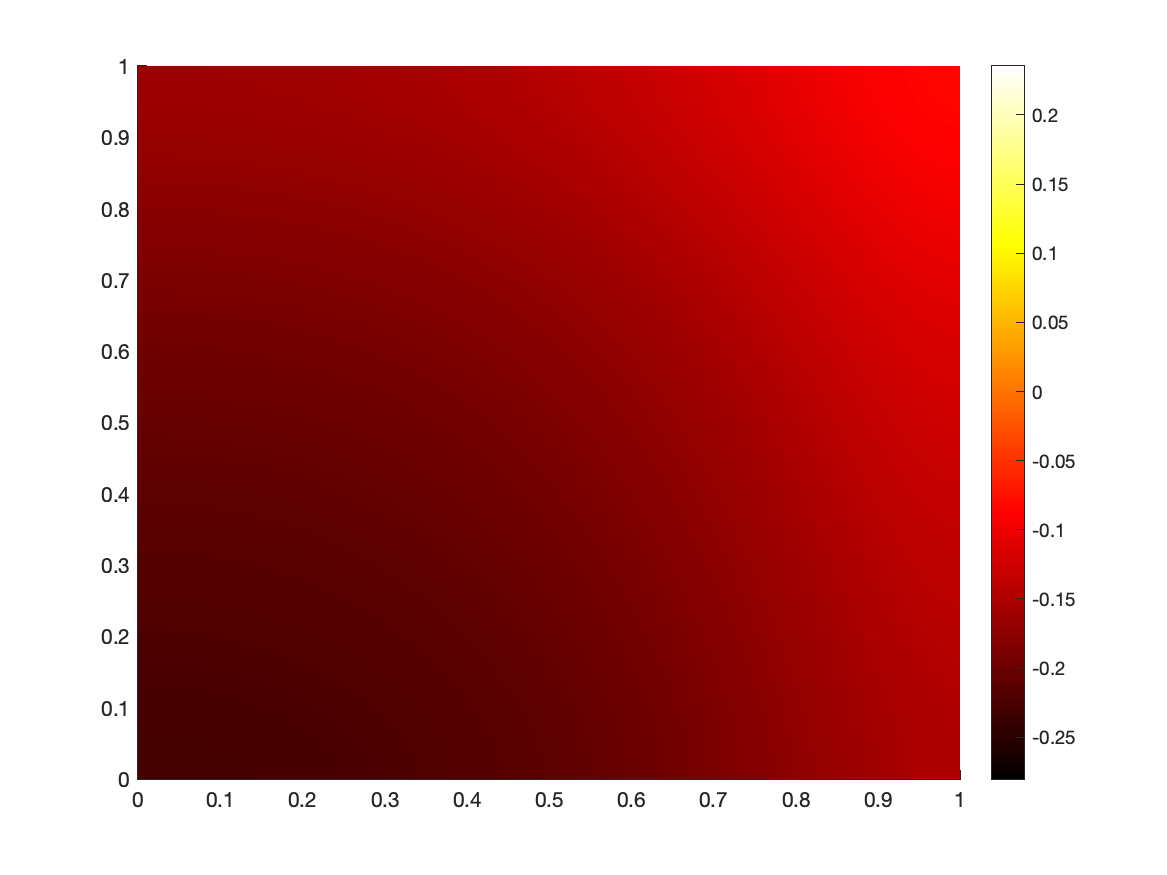}}
     \subfloat[PBDW $\eta_{2,50}$]{
        \includegraphics[width=0.25\textwidth]{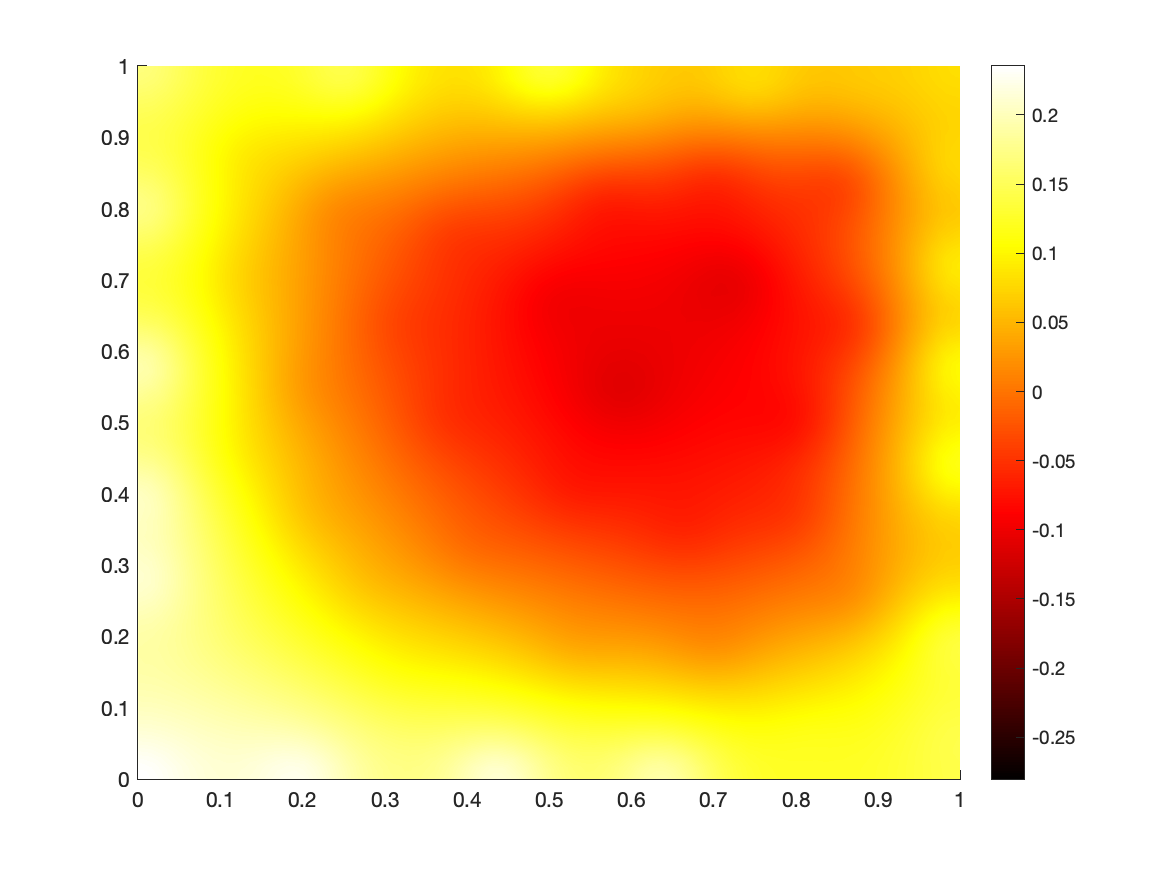}}
    \subfloat[PBDW $u_{2,50}$]{
        \includegraphics[width=0.25\textwidth]{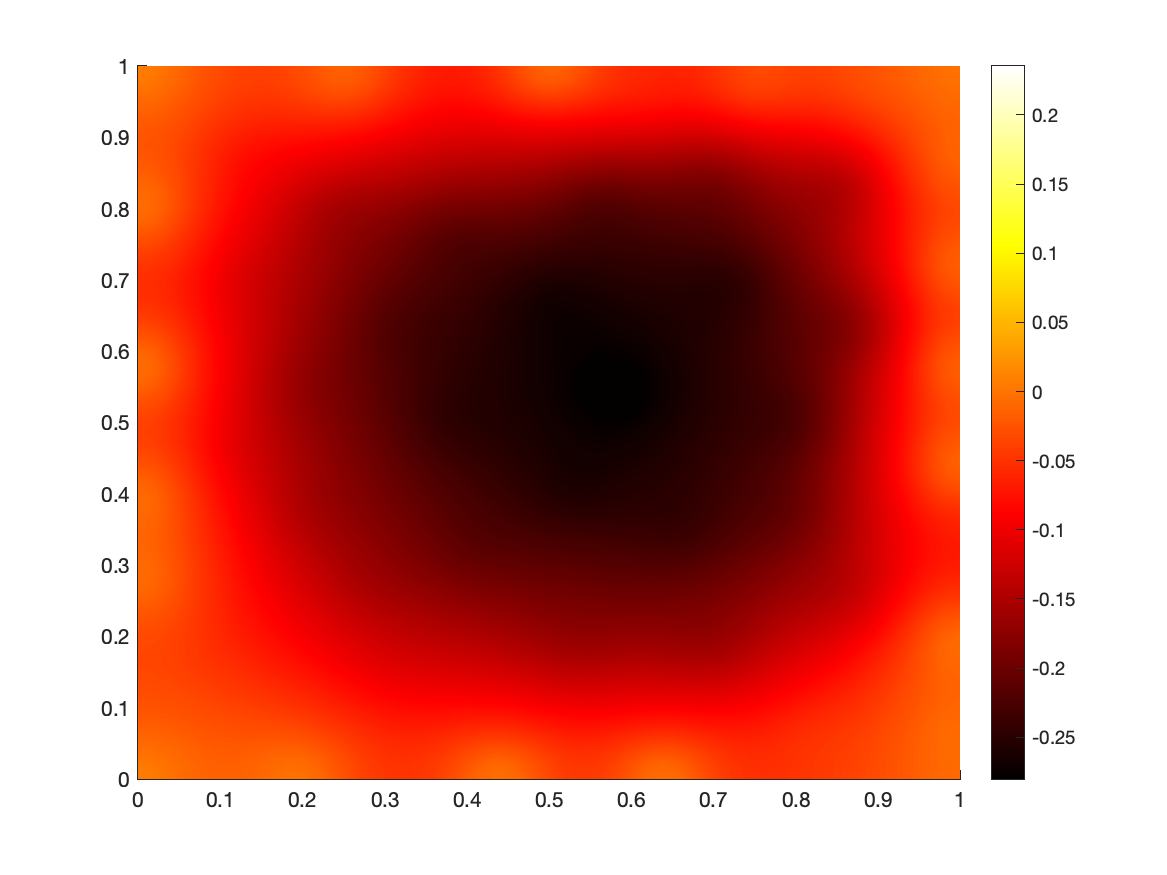}}
    \subfloat[$\left |u_{\text {true }}-u_{2, 50}\right |$]{
        \includegraphics[width=0.25\textwidth]{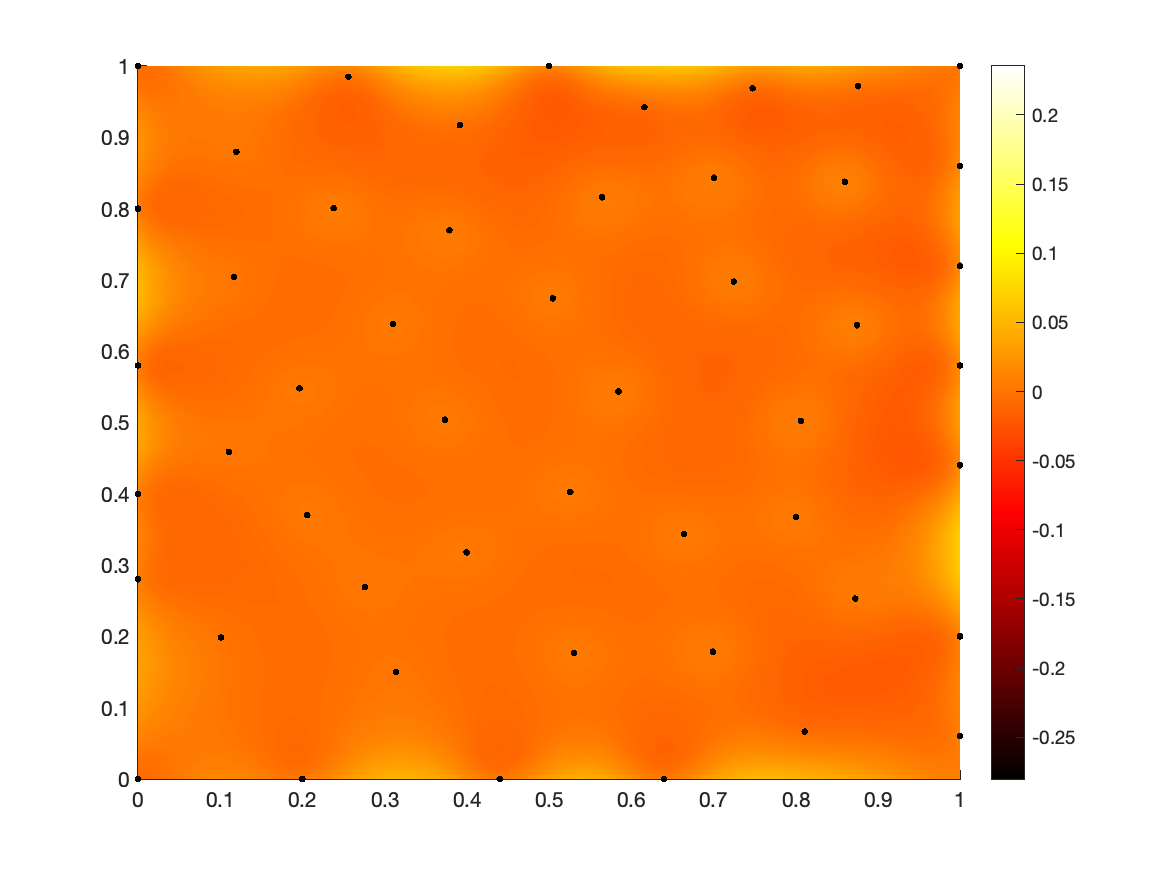}}

    \vspace{0.4cm}

    \subfloat[PBDW $z_{6,50}$]{
        \includegraphics[width=0.25\textwidth]{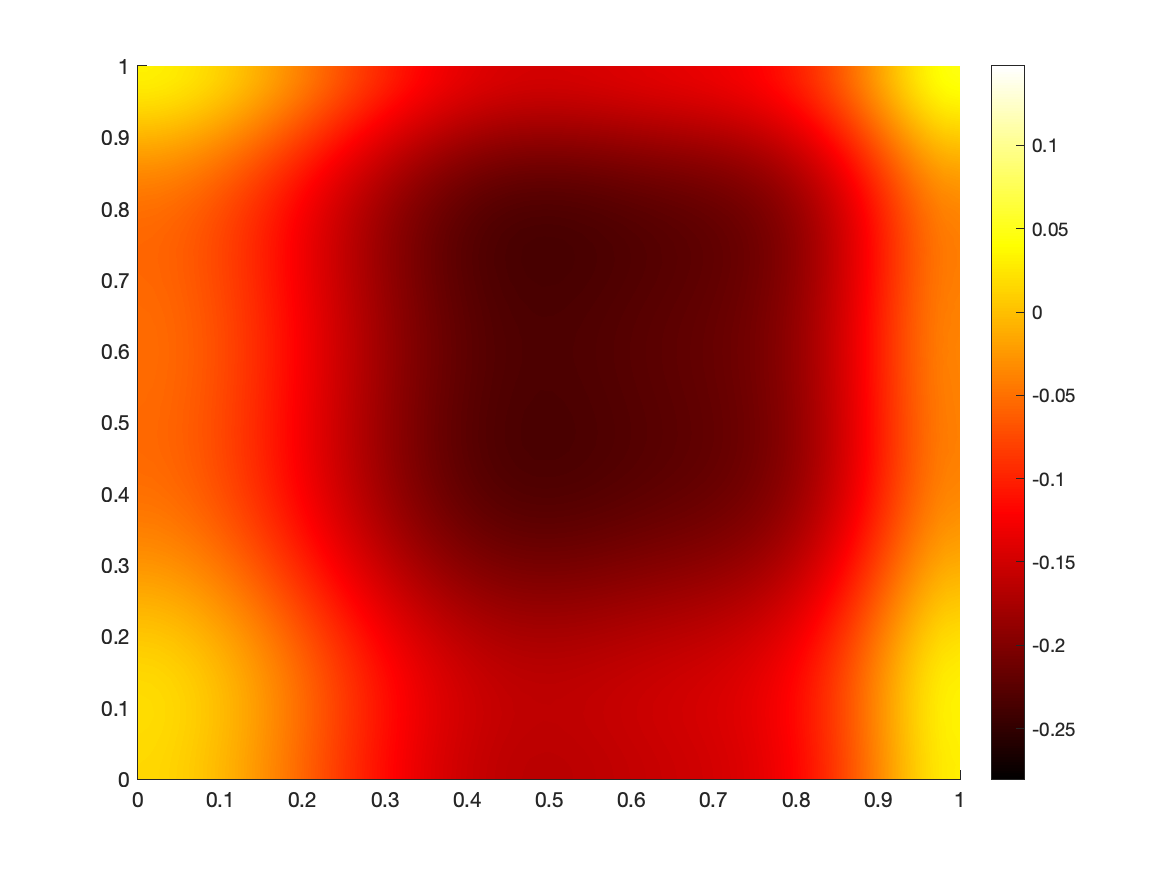}}
     \subfloat[PBDW $\eta_{6,50}$]{
        \includegraphics[width=0.25\textwidth]{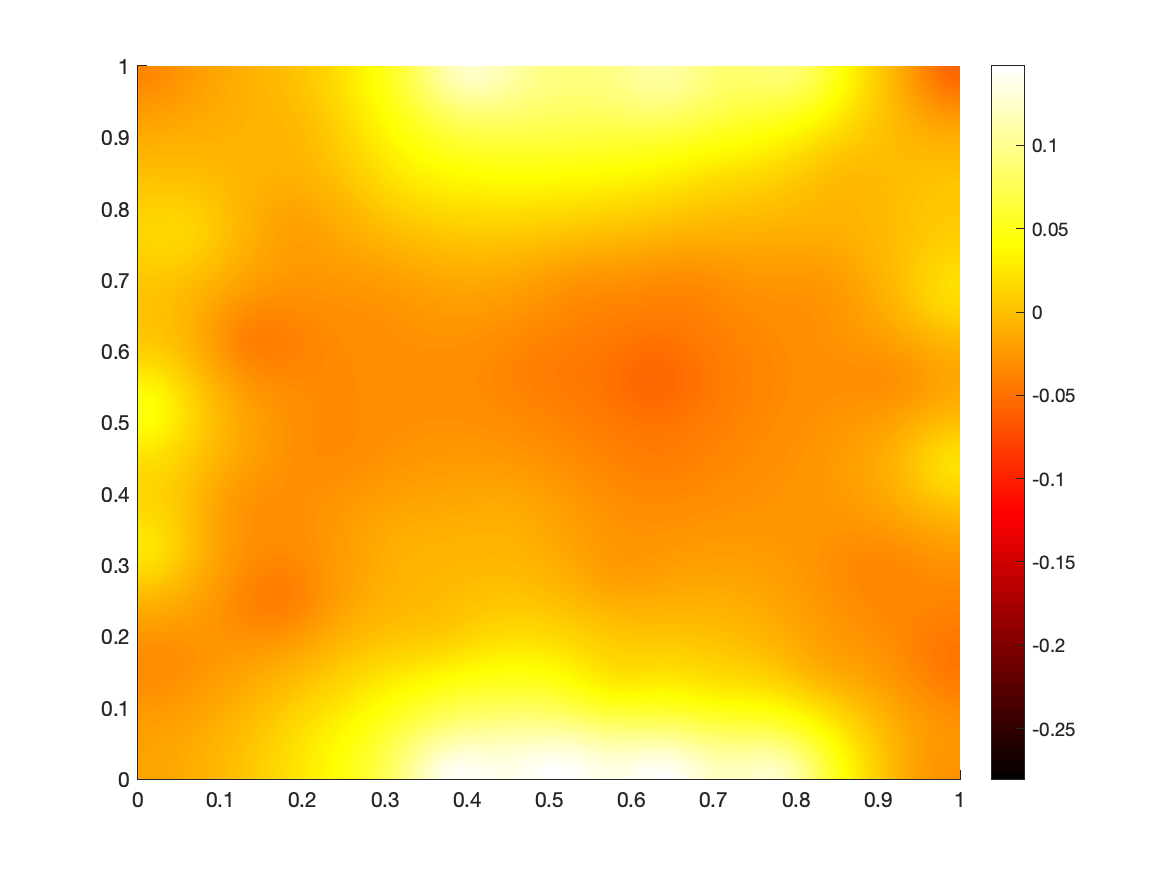}}
    \subfloat[PBDW $u_{6,50}$]{
        \includegraphics[width=0.25\textwidth]{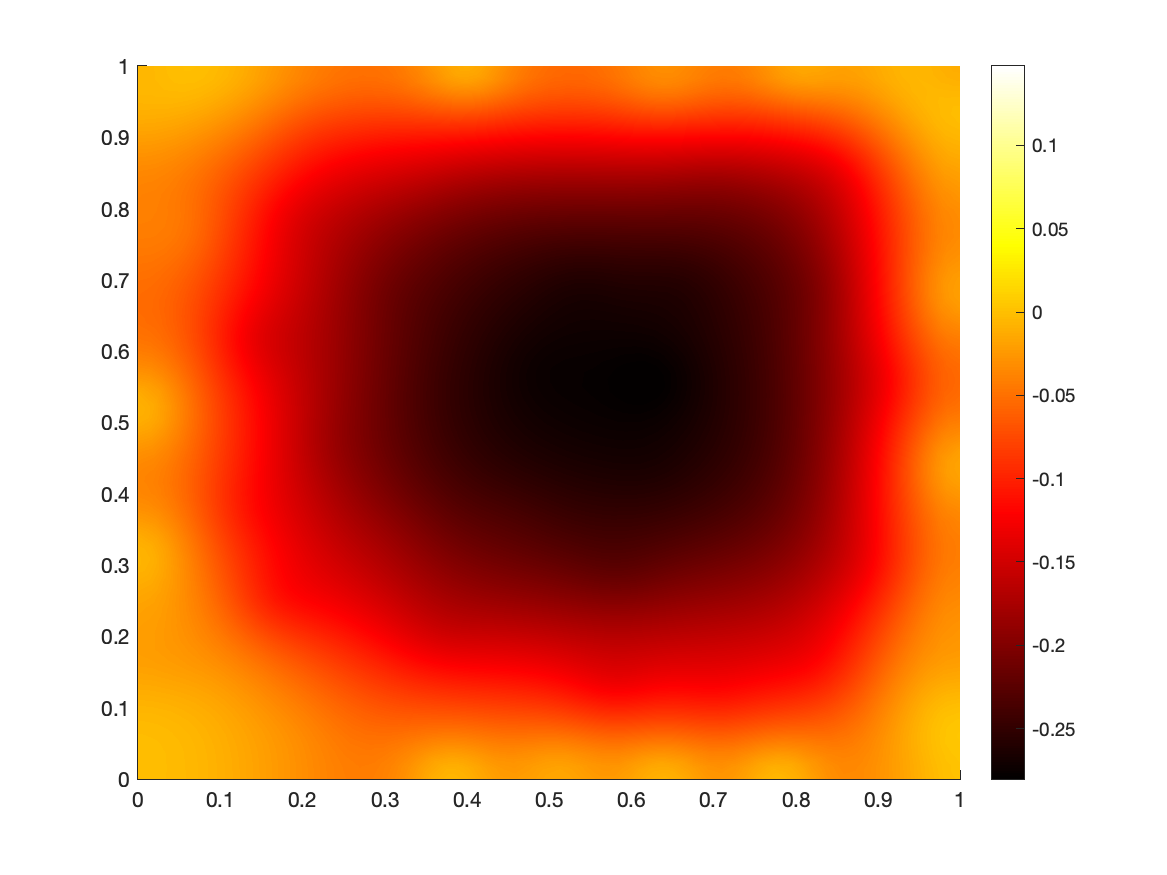}}
    \subfloat[$\left |u_{\text {true }}-u_{6, 50}\right |$]{
        \includegraphics[width=0.25\textwidth]{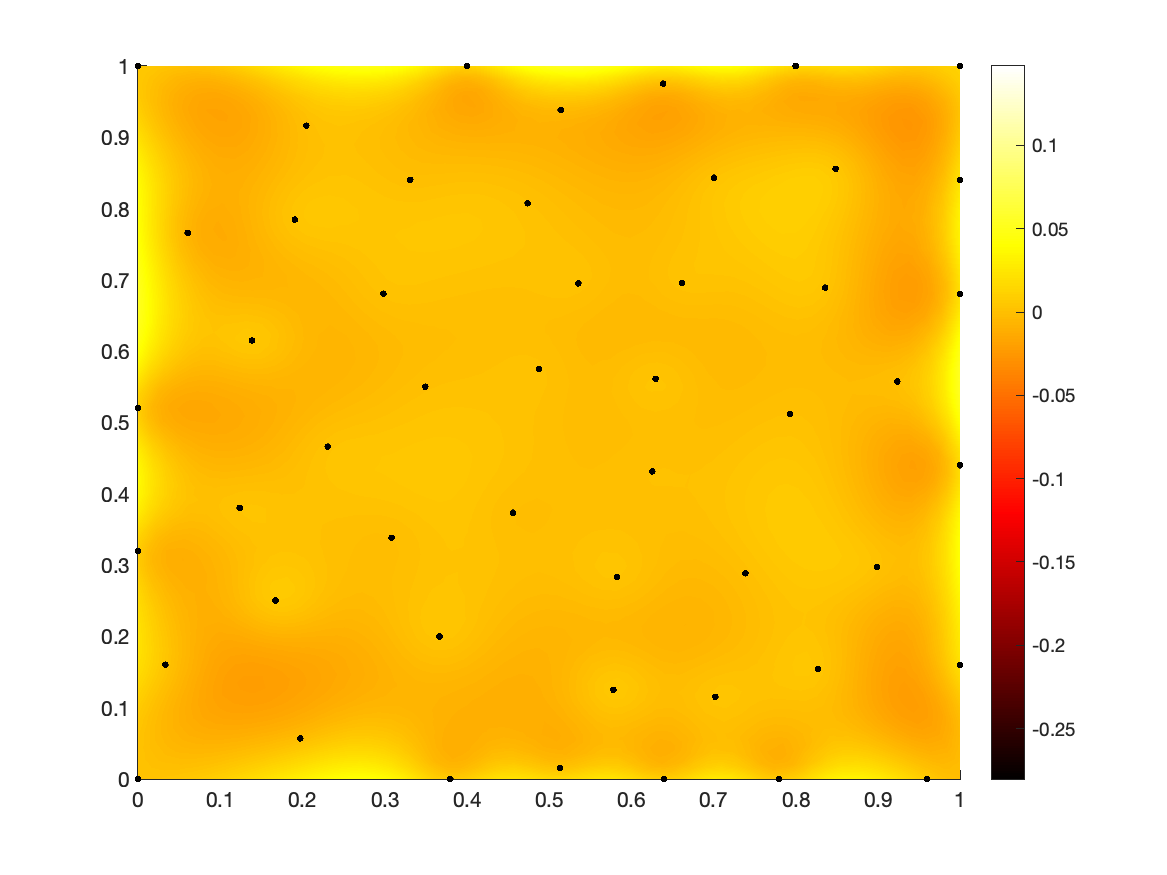}}

    \vspace{0.4cm}

    \subfloat[PBDW of $z_{15,50}$]{
        \includegraphics[width=0.25\textwidth]{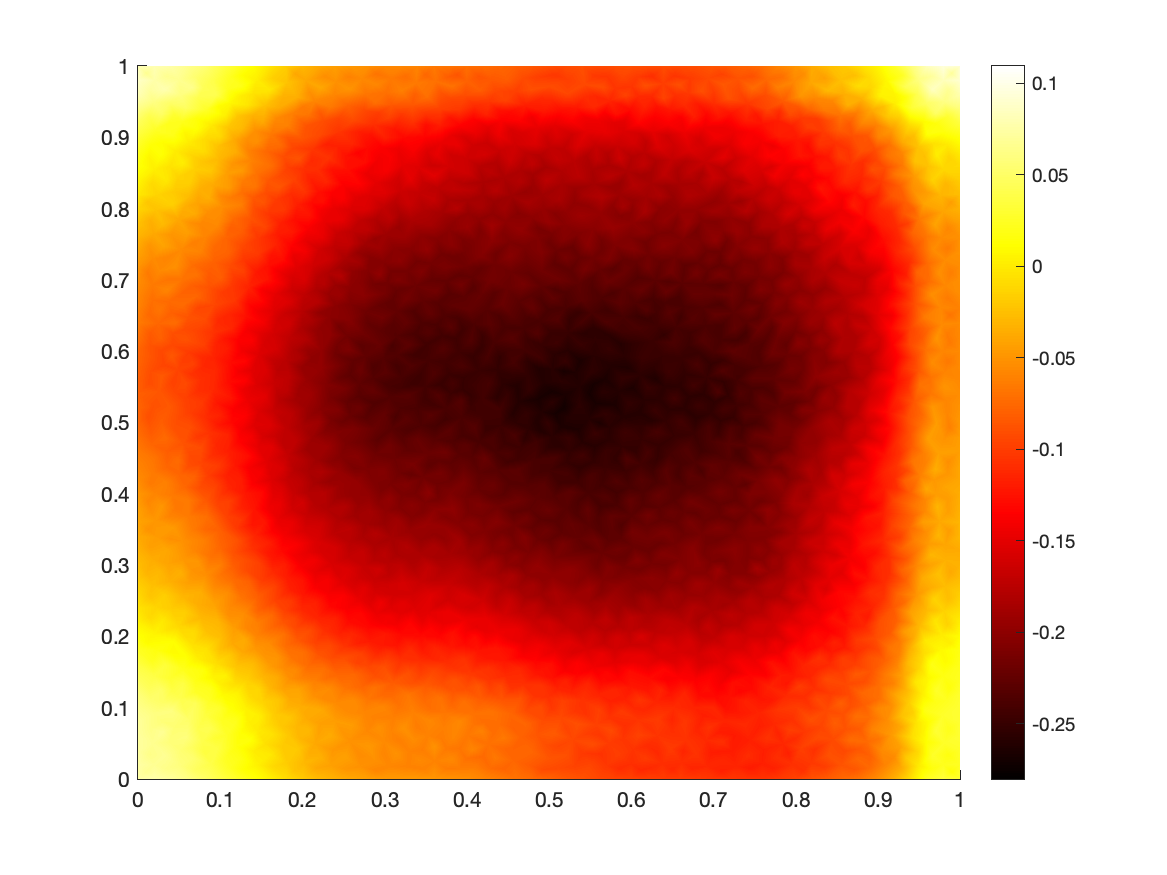}}
     \subfloat[PBDW $\eta_{15,50}$]{
        \includegraphics[width=0.25\textwidth]{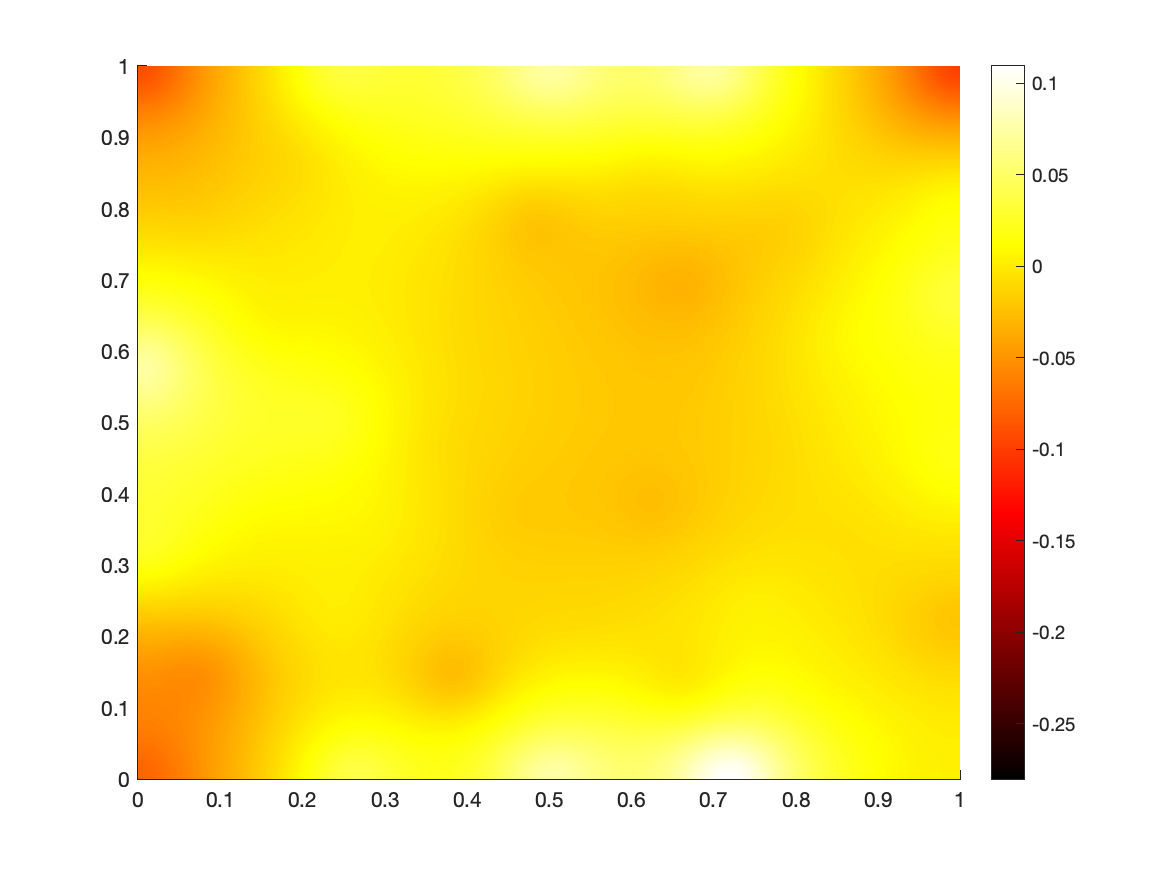}}
    \subfloat[PBDW $u_{15,50}$]{
        \includegraphics[width=0.25\textwidth]{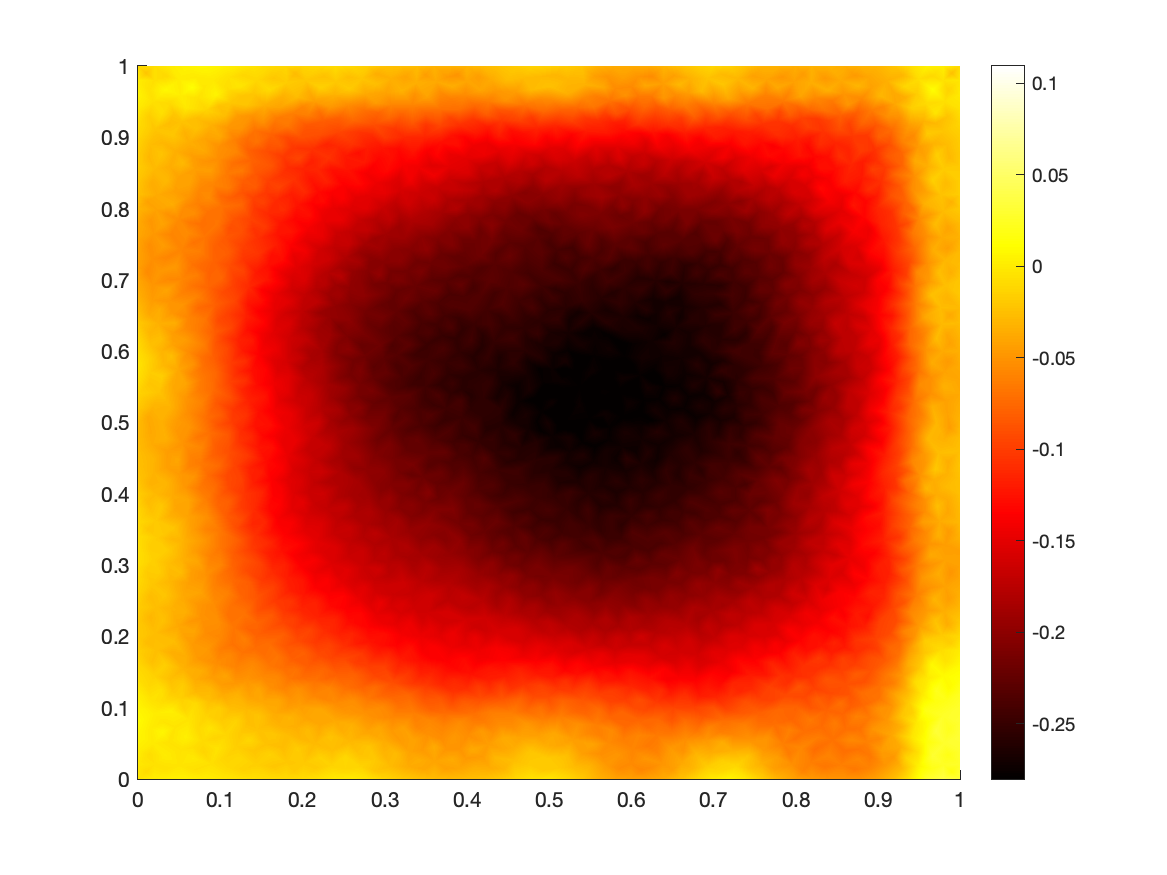}}
    \subfloat[$\left |u_{\text {true }}-u_{15, 50}\right |$]{
        \includegraphics[width=0.25\textwidth]{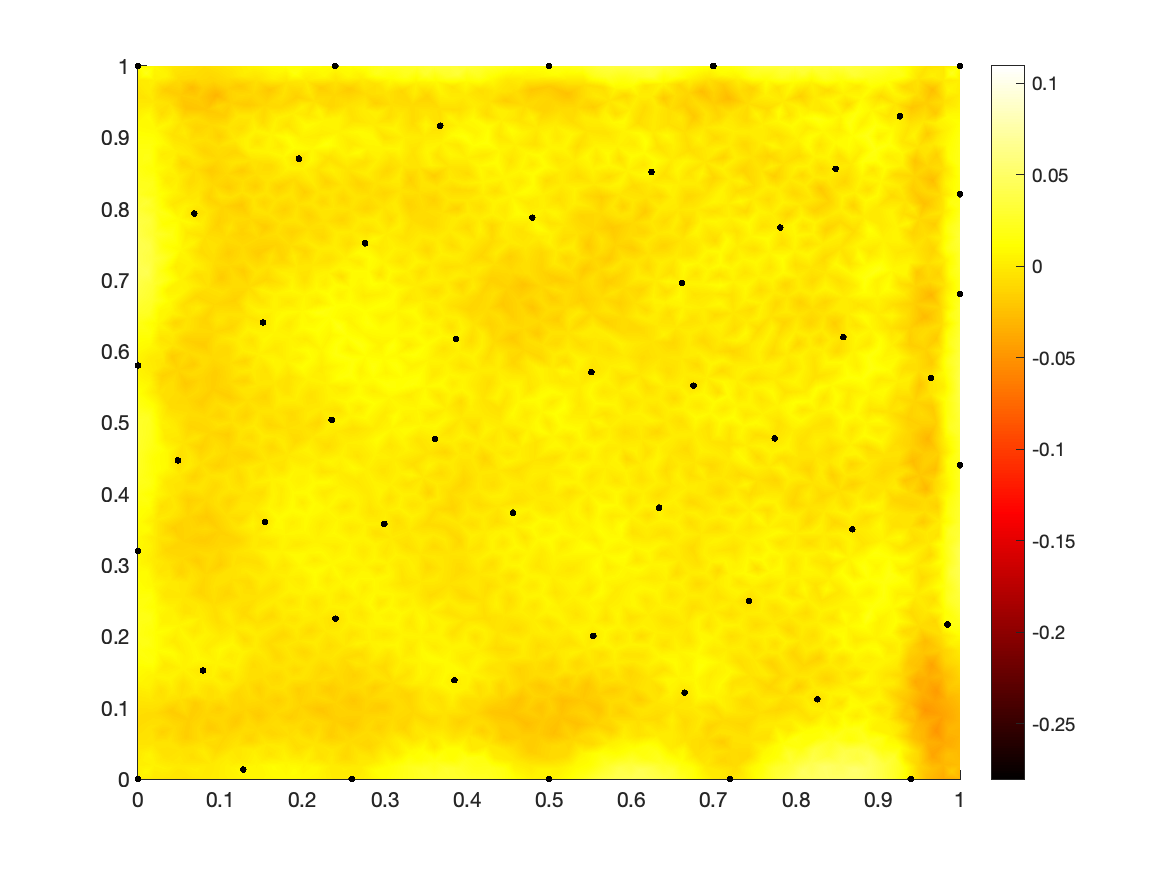}}

    \caption{Evolution of fields as a function of the number of modes in the background space for a perfect model without measurement noise for M = 50 sensors and different number of reduced order modes N = {2,6,15}.}
    \label{fig:local_errors}
\end{figure*}

The quantities of interest introduced above are shown in Figure~\ref{fig:global_errors}. Their evolution closely follows that of the projection error. In the case of a perfect model, the dominant source of error arises from the approximation of the true state within the background space. As this space is enriched (that is, as $N$ increases), the accuracy of the reconstruction improves accordingly, leading to a  reduction in the total error.\\

For small background dimensions, specifically when only the first and second modes are considered, the PBDW formulation exhibits superior performance compared to a pure reduced-basis approach. The projection error of the reduced basis becomes smaller than the reconstruction error for $N > 2$, indicating that background spaces of dimension $N \leq 2$ provide an insufficient model representation. In this regime, the correction term $\eta$  plays a crucial role in compensating for the missing components of the state that lie outside the background space.\\

Figure~\ref{fig:local_errors} highlights another effect related to the choice of the reduced dimension $N$. As the background space is enriched(for increasing values of $N$), the error field tends to concentrate in regions that lie far from the sensor locations, where the available information is not sufficient to fully constrain the reconstruction.

\subsection{Biased model without regularization}

We  now simulate the impact of different biases on the recovery task. As a reminder, the parametric model is replaced by the background approximation $z $.  The model error or model bias $\eta$ is not anticipated by the background approximation space, and therefore does not belong to this first approximation space, but to the update space used to correct the model estimate with information from measurements. In the following study we are considering $N=2$ background modes.


\begin{figure*}[htp]
    \centering

    \subfloat[Prediction of $\eta_{2,50}$]{
        \includegraphics[width=0.32\textwidth]{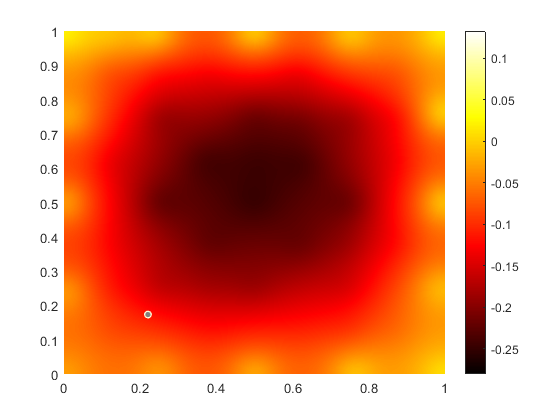}
    }
    \subfloat[PBDW $\eta_{2,50}$]{
        \includegraphics[width=0.32\textwidth]{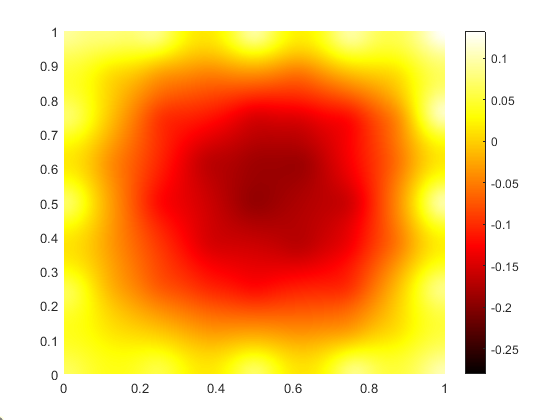}
    }
    \subfloat[PBDW $z_{2,50}$]{
        \includegraphics[width=0.32\textwidth]{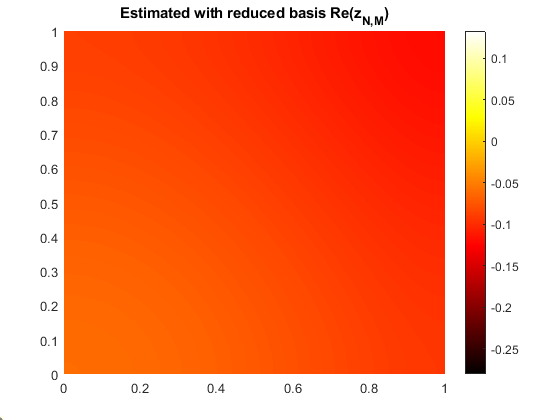}
    }

    \vspace{0.4cm}

    \subfloat[Hybrid twin $u_{2,50}$]{
        \includegraphics[width=0.32\textwidth]{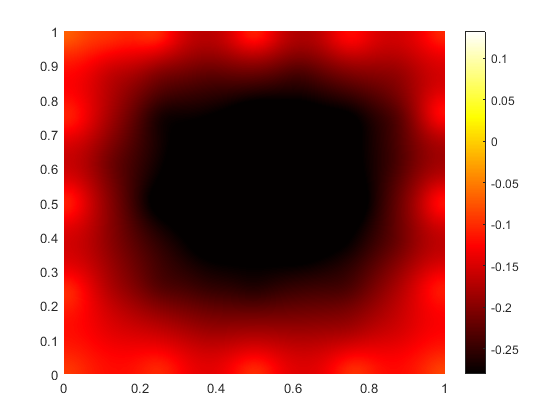}
    }
    \subfloat[PBDW $u_{2,50}$]{
        \includegraphics[width=0.32\textwidth]{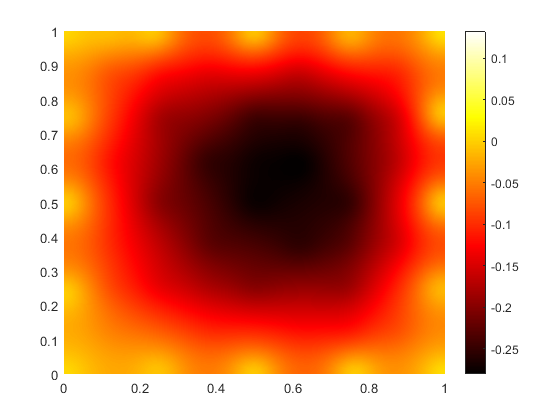}
    }
    \subfloat[Ground truth $u$]{
        \includegraphics[width=0.32\textwidth]{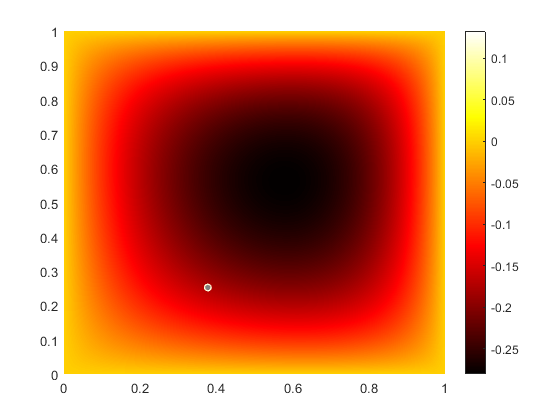}
    }

    \caption{Simulation with perfect observations and misspecified boundary conditions under strong orthogonality.}
    \label{fig:nbRayons}
\end{figure*}

\begin{figure*}[htp]
    \centering

    \subfloat[Prediction of $\eta_{2,50}$]{
        \includegraphics[width=0.32\textwidth]{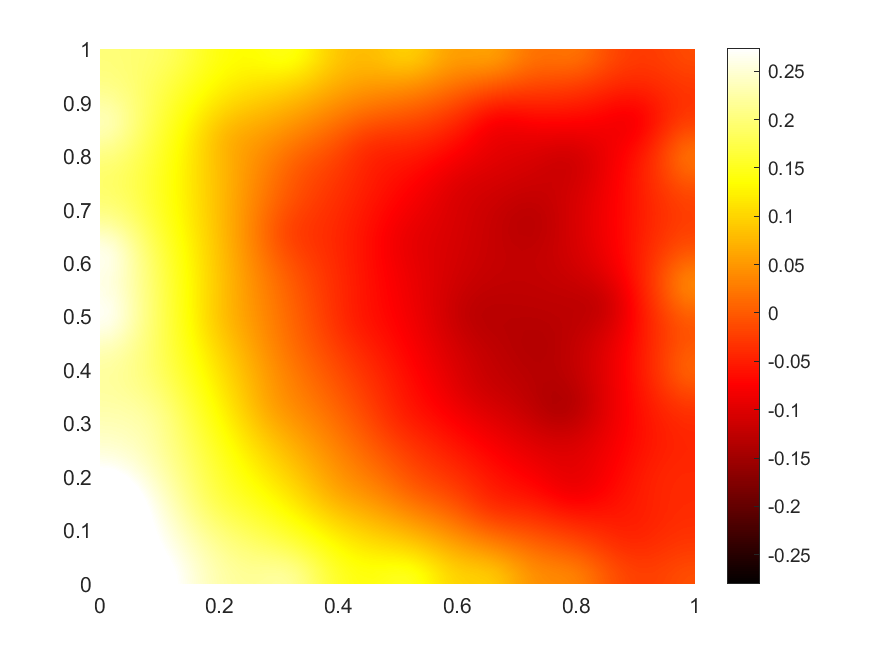}
    }
    \subfloat[PBDW $\eta_{2,50}$]{
        \includegraphics[width=0.32\textwidth]{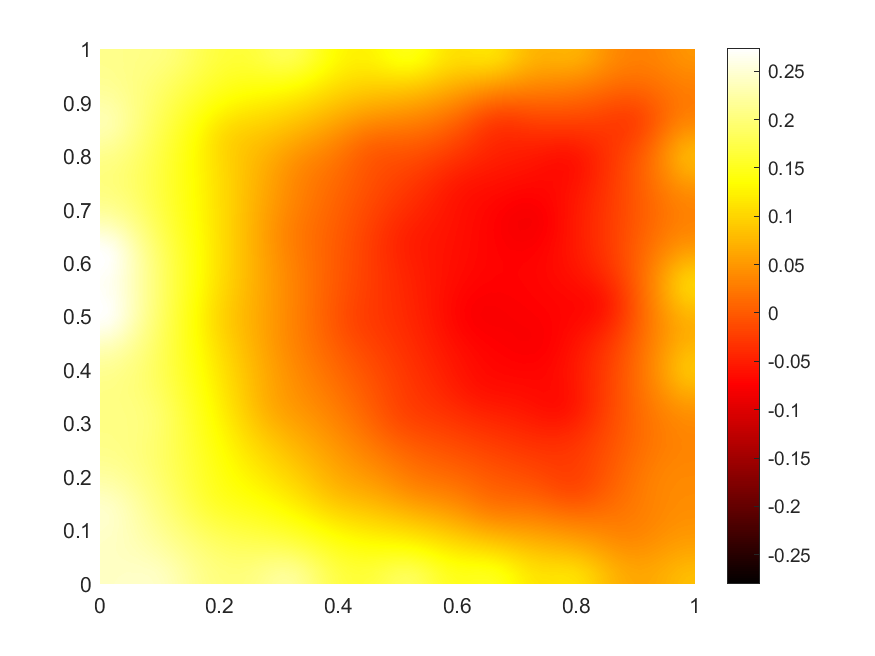}
    }
    \subfloat[PBDW $z_{2,50}$]{
        \includegraphics[width=0.32\textwidth]{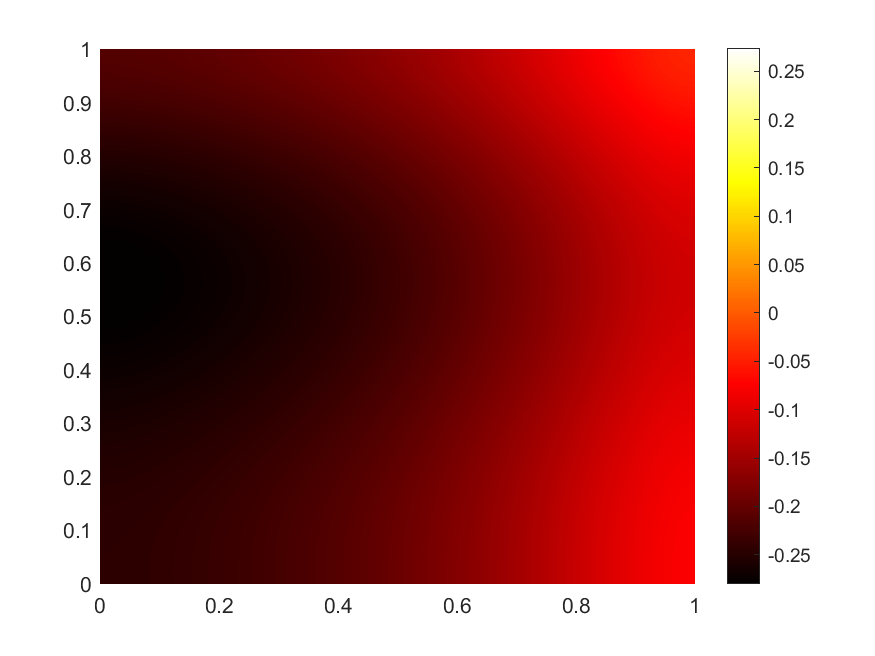}
    }

    \vspace{0.4cm}

    \subfloat[Hybrid twin $u_{2,50}$]{
        \includegraphics[width=0.32\textwidth]{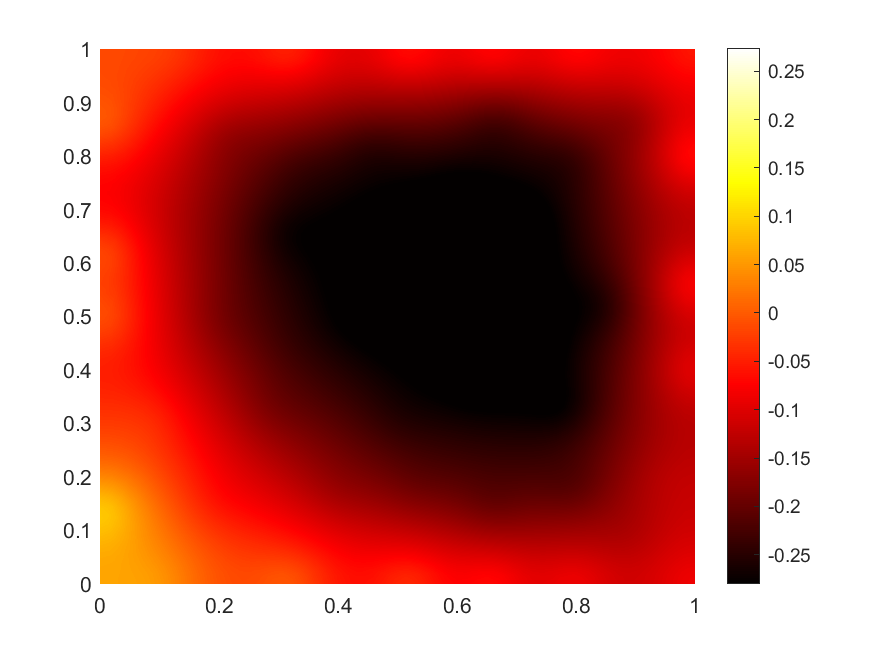}
    }
    \subfloat[PBDW $u_{2,50}$]{
        \includegraphics[width=0.32\textwidth]{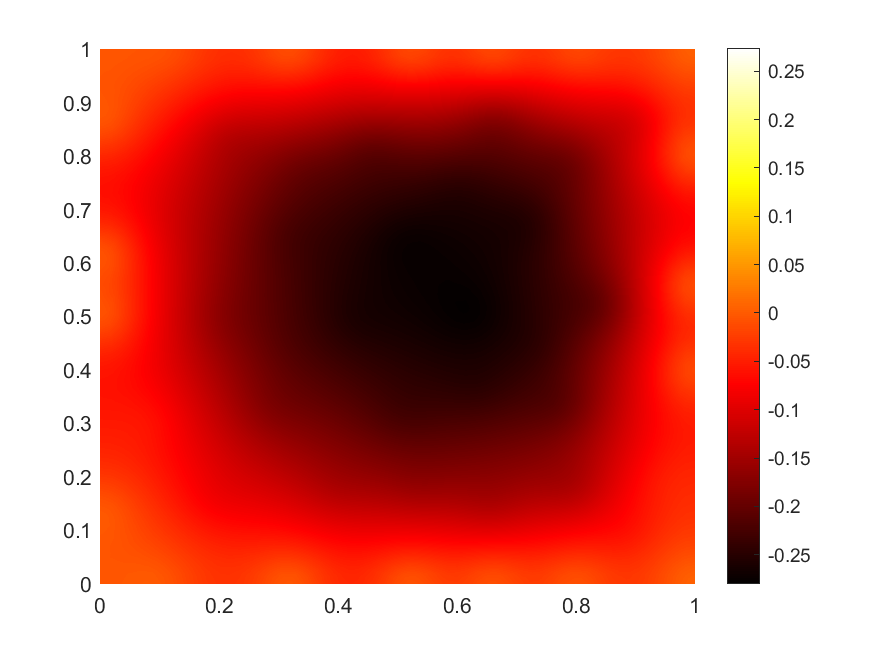}
    }
    \subfloat[Ground truth $u$]{
        \includegraphics[width=0.32\textwidth]{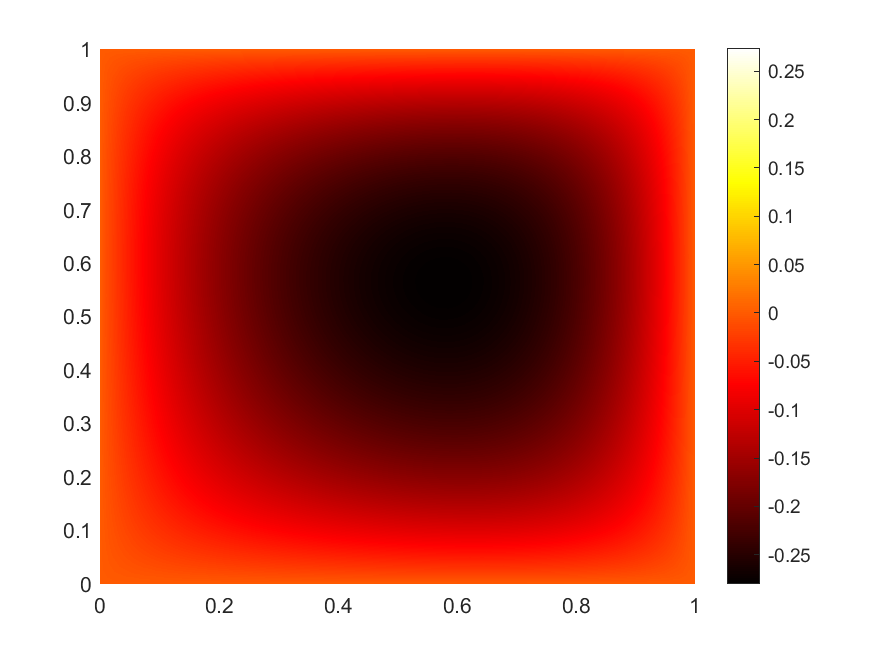}
    }

    \caption{Simulation with perfect observations, misspecified boundary conditions, and misspecified source under strong orthogonality.}
    \label{fig:nbRayons2}
\end{figure*}

In both Figures~\ref{fig:nbRayons} and ~\ref{fig:nbRayons2}, (a) depicts the predicted update $\eta_{\theta}$ (model-based correction) produced by our method; (b) shows the estimated update $\eta$ (data-driven correction) obtained via PBDW; (c) presents the background estimation $z$ computed by PBDW; (d) illustrates the full-state reconstruction $z + \eta_{\theta}$ generated by the proposed Hybrid Twin; (e) shows the full-state reconstruction $z + \eta$ obtained by PBDW; and (f) displays the ground truth $u^{\text{true}}$, approximated using the Finite Element Method. For $N = 2$ reduced modes, the background estimation $z$ in both cases deviates considerably from the reference $u^{\text{true}}$. The corrective components ($\eta_{\theta}$ and $\eta$) therefore play a critical role in incorporating sensor information to enhance the reconstruction accuracy, particularly when the reduced basis dimension $N$ is small.

\begin{figure*}[htp]
    \centering

    \subfloat[PBDW]{
        \includegraphics[width=0.47\textwidth]{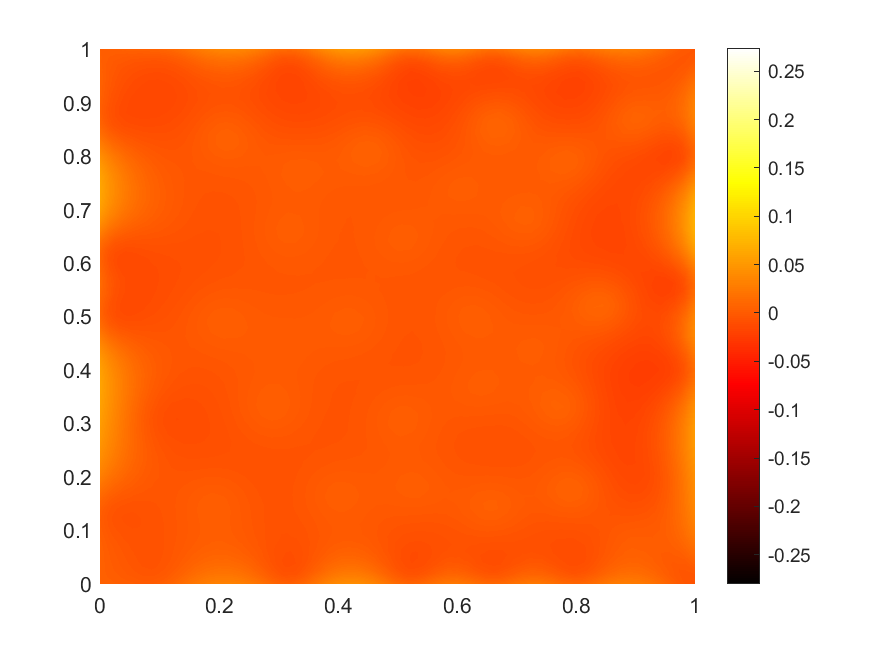}
    }
    \subfloat[Hybrid twin]{
        \includegraphics[width=0.47\textwidth]{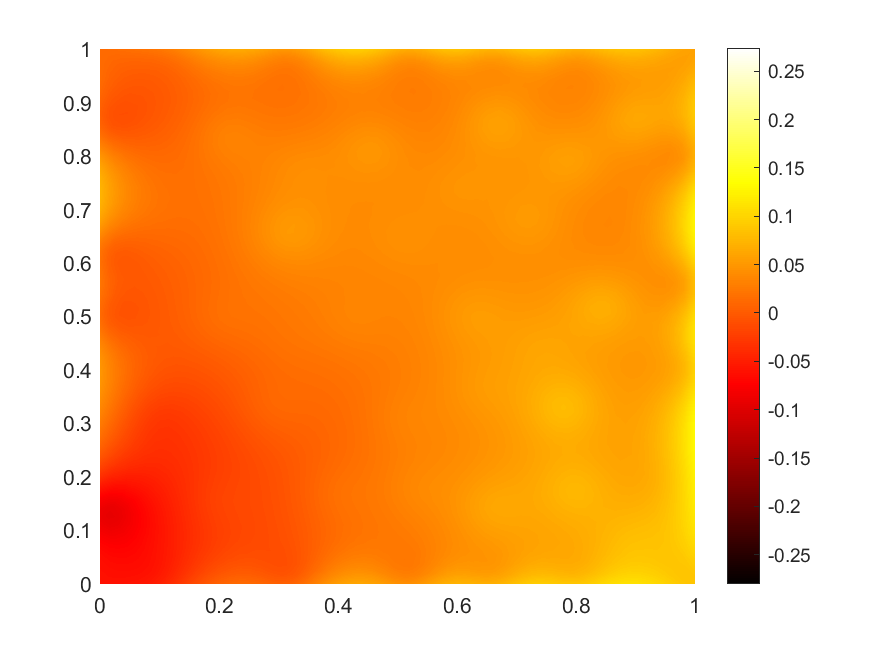}
    }

    \caption{Relative $L_{2}(\Omega)$ error.}
    \label{fig:nbRayons3}
\end{figure*}

\subsection{Impact of the regularization weight $\xi$ }

We now consider the case of noisy measurements.  
Let the observational noise be homoscedastic, uncorrelated, and of zero mean:
\begin{equation}
    \mathbb{E}\bigl[\,\delta_m\,\bigr] = 0, \qquad m = 1, \dots, M.
\end{equation}

The relative noise level is defined as
\begin{equation}
    \delta_m = \frac{\| y_m^\delta - y_m \|_{L^2}}{\| y_m \|_{L^2}}
\end{equation}
The noisy observations are modeled as:
\begin{equation}
    y_m^\delta = y_m \,\bigl( 1 + \delta \, r_m \bigr).
\end{equation}
Where $\|\cdot\|_{L^2}$ denotes the $L^2$-norm and $r_m$ represents the prescribed standard deviation of the noise. For clarity of terminology, we refer to the non-regularized formulation as \textbf{PBDW}, and to its regularized counterpart (Equation~\eqref{eq:pbdw_regul}) as \textbf{APBDW}.

\begin{figure}[H]
    \centering
    \includegraphics[width=0.55\linewidth]{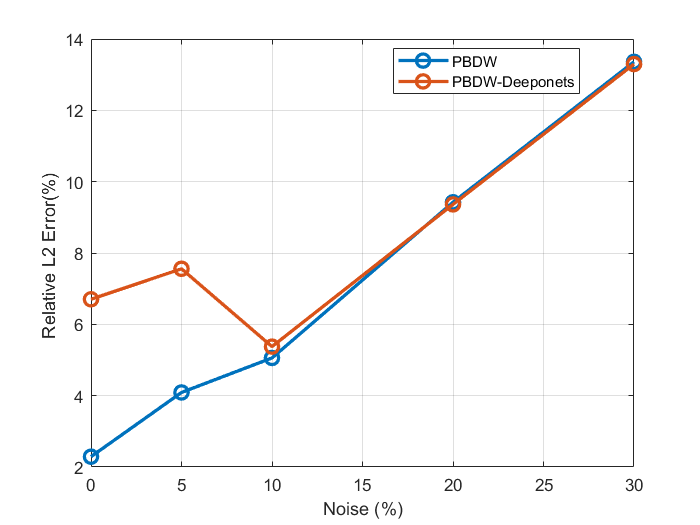}
    \caption{Evolution of the relative $L^2$ error as a function of measurement noise.}
    \label{fig:noise_apbdw}
\end{figure}

Both the original PBDW and the PBDW-DeepONet exhibit robustness to noise levels up to 10\%. Beyond this threshold, the relative $L^2$ error increases approximately linearly with the noise amplitude. The PBDW is a projection-by-data framework, meaning that if $N$ denotes the dimension of the background space and $M \gg N$ the dimension of the update space, the method's performance becomes highly sensitive to the noise present in the $M$ observations. This mutual convergence behavior can be explained by the fact that the model-based estimation is identical for both methods; thus, variations in the relative error are primarily attributed to the corrective updates, whose contributions remain small. In the following, we evaluate both the standard PBDW and the regularized variant designed to mitigate the effect of noise on state estimation.

The regularization parameter controls the balance between the contributions of the background space and the observation space when the measurements are noisy. In this study, the optimal regularization parameter is selected using the generalized cross-validation (GCV) method for each noise level. Figure~\ref{fig:noise_apbdw} shows the evolution of the relative $L^2$ error as a function of the noise level in the observations. The impact of regularization is clearly observed: the two regularized methods (APBDW and APBDW-DeepONet) achieve superior performance under high noise levels, whereas the non-regularized approaches (PBDW and PBDW-DeepONet) lose accuracy. The regularization mechanism increasingly prioritizes the model as the observation noise grows, compensating for the degradation of data quality. Consequently, the relative error does not necessarily increase with the noise amplitude. Furthermore, as the noise level increases, the APBDW-DeepONet converges toward the performance of APBDW, and similarly, PBDW-DeepONet converges toward PBDW.

\begin{figure*}[htp]
    \centering

    \subfloat[Randomly selected sensors]{
        \includegraphics[width=0.47\textwidth]{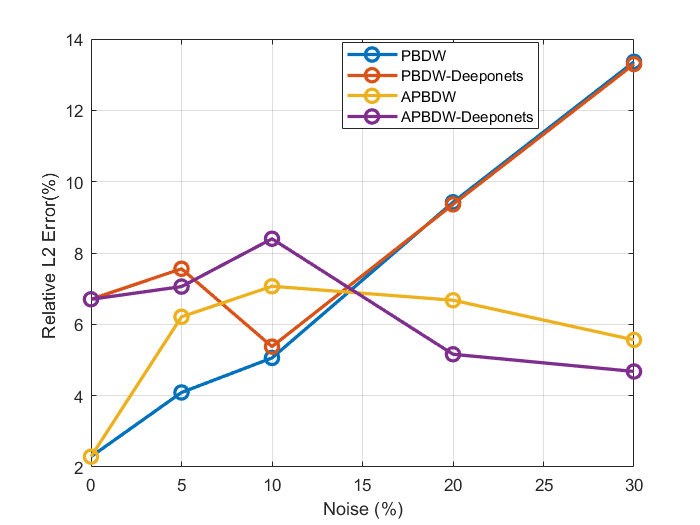}
    }
    \subfloat[Optimally selected sensors]{
        \includegraphics[width=0.47\textwidth]{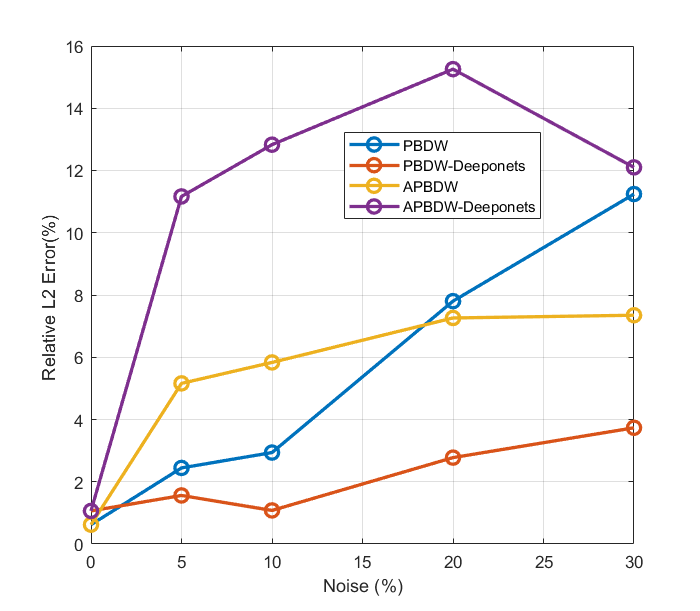}
    }

    \caption{Influence of measurement noise on the $L^2$ error for different sensor selection strategies.}
    \label{fig:dernier}
\end{figure*}

\begin{figure}[H]
    \centering
    \includegraphics[width=0.85\linewidth]{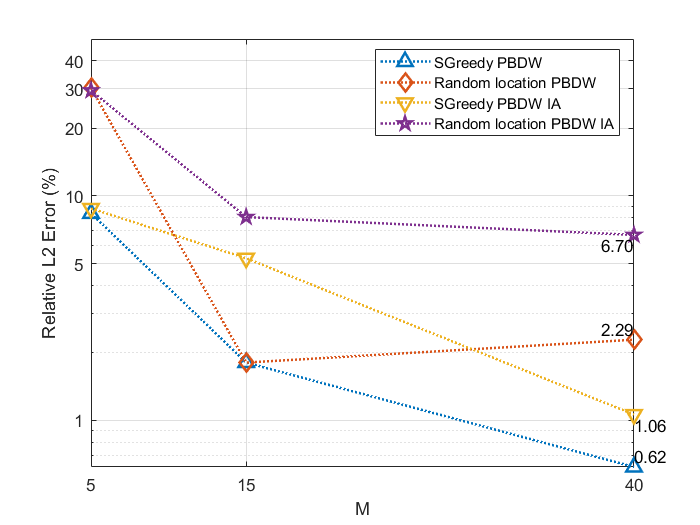}
    \caption{Evolution of the relative $L^2$ error as a function of measurement noise for various sensor selection strategies. }
    \label{fig:number_sensors}
\end{figure}

To describe at best the update space, here the input function of the DeepONet is discretized at sensors locations that are optimally selected with the stability maximization algorithm. Figure \ref{fig:dernier} shows the influence of the observation noise on the relative error of the four approaches and exhibits how the choice of the sensors location affects the system behavior. The noise level of interest here are from 0\% to 30\%. The frameworks without regularization (PBDW, PBDW-DeepONet) show better performance when the sensors are selected through the SGreedy algorithm whatever the noise level.\\

On the other hand, the regularized frameworks behave differently. For (APBDW-DeepONet, APBDW) the selection of sensors through SGreedy has better performance for low noise level. As the noise level increases until certain point, the random selection of sensors outperform the SGreedy procedure. That point is reached almost immediately (3\%) when the update is computed by the neural operator (APBDW-DeepONet) and around 16\% for the classical APBDW without AI. Out of the four frameworks, the PBDW-DeepONet is the most robust to noisy measurements. It achieves 3.7\% relative error in the state estimation even with 30\% of observation noise. Figure \ref{fig:number_sensors} shows how the number of sensors and their location choice (SGreedy or random) affect the quality of the estimation.\\

We see that the relative error globally decreases with the number of sensors and stability-maximization(Sgreedy) procedure provides a better state estimate than a random selection of sensors. 
The Sgreedy selection of sensors outperforms the random selection by far especially when $\frac{M}{N} \to 1$. 

\subsection{Limitations }
\begin{figure}[H]
    \centering
    \includegraphics[width=0.65\linewidth]{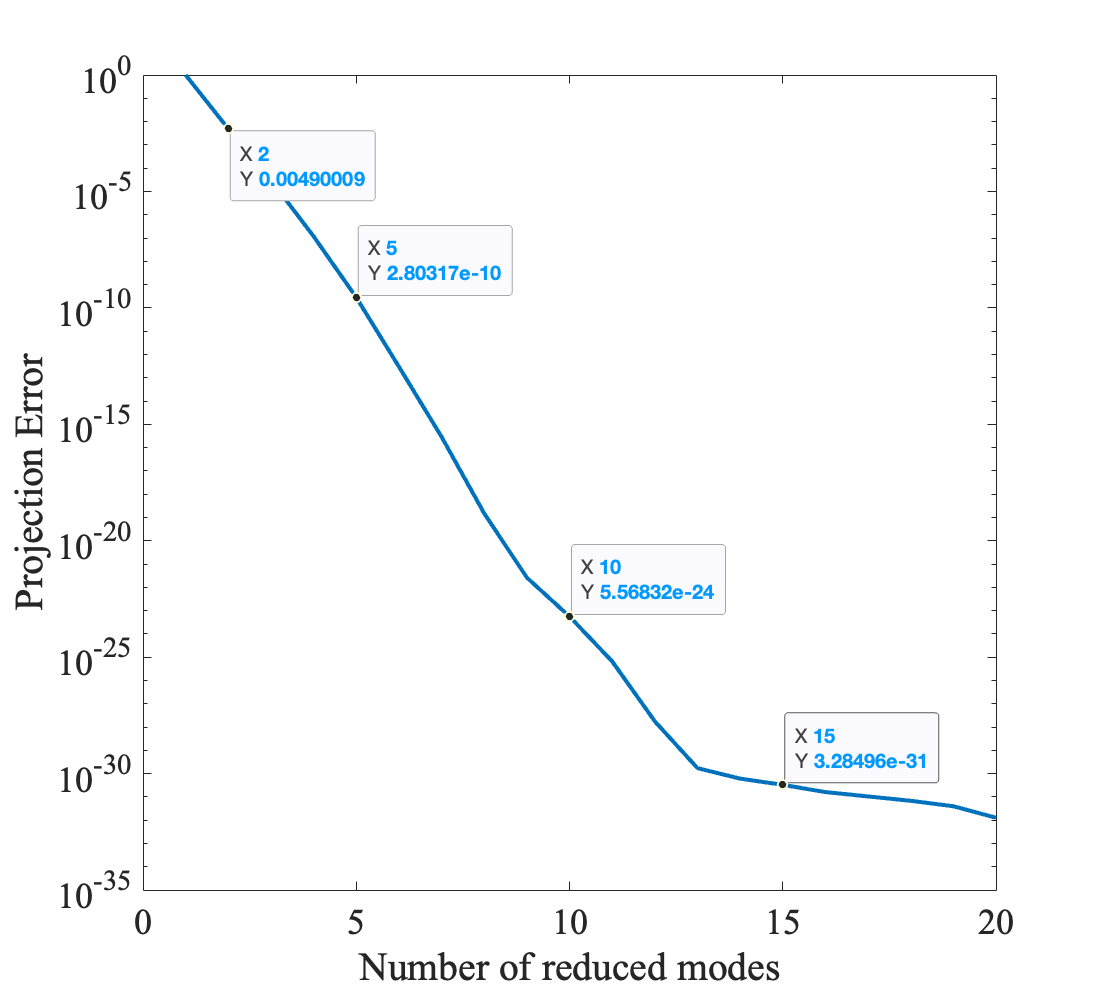}
    \caption{Reductibility of the best knowledge manifold $\mathcal{M}_{bk}$.}
    \label{fig:bases_reduites}
\end{figure}

Both the classical PBDW and the combined PBDW-DeepONets are computed in two steps. The  construction of a background space and an observation space made offline, and an online step for the inversion. The training of the DeepONet is computed offline for PBDW-DeepONets. The best knowledge manifold $\mathcal{M}_{bk}$ of the Helmholtz equation is reductible, meaning the background space can approximate $M_{bk}$  with small value of $N$. Figure~\ref{fig:bases_reduites} shows that $\mathcal{Z}_N$'s approximation error reaches $5\cdot10^{-3}$ with only $N=2$. 
Therefore the computational time is mainly influenced by $M$, as \( N \ll M \), in this paper \(M = 25N\).

In PBDW-DeepONets, computing the update $\eta$ requires matrix-vector multiplications. 
On a MacBook M3 Pro with $M=50$, the classical PBDW update estimation $\eta$ is obtained in approximately 10 milliseconds on the CPU, whereas the DeepONets-based update implemented in \textbf{Numpy} takes around 20 milliseconds, i.e., 10 milliseconds longer. 
This computational time can be significantly reduced by employing optimized frameworks such as \textbf{Jax}. 
Moreover, on a GPU, the parallel execution of the matrix-vector operations allows for even faster inference. 
As a result, the PBDW-DeepONets approach is compatible with real-time applications, depending on the temporal characteristics of the problem.

\begin{table}[h!]
\centering
\caption{Comparison of practical computational cost between PBDW and PBDW-DeepONet for $M=50$.}
\label{tab:timing_comparison}
\begin{tabular}{|l|l|}
\hline
\textbf{Method} & \textbf{Online Time} \\
\hline
PBDW $\eta$ estimation & 10 ms (CPU) \\
\hline
PBDW-DeepONets $\eta$ prediction & 20 ms (CPU) \\

\hline
\end{tabular}
\end{table}

\section{Conclusion}

We demonstrated that physics-constrained optimization can be hybridized with machine learning for a data assimilation task. In this paper, we coupled the variational data assimilation framework PBDW with DeepONet to estimate the state of system while predicting the unanticipated uncertainties (ignorance, bias). The bias-aware hybrid AI was implemented using two approaches to ensure orthogonality between the background (what is known) and the bias (what is learned).\\

The first approach uses a weak constraint, which penalizes the loss function for an orthogonality violation defined by an inner product. In this case, the trunk network is a neural network (as introduced in the original DeepONet) that learns spatial basis functions defined globally, allowing inference at any location.
The second approach involves a strong constraint, where the trunk is replaced by a set of specific basis functions, enabling optimal inference at a fixed sensors locations optimally selected.\\

While both methods demonstrated various strengths and weaknesses, the numerical results showed that the strong constraint outperforms the weak constraint in terms of the convergence of the loss function. Additionally, it guarantees the orthogonality property during both the training and inference stages. The current hybrid-AI was tested for a given model's parameter value (wave number) with different forcing functions. However, the proposed approach is flexible to include the model's parameter as an additional input.

\section*{Declarations}


\subsection*{Funding} 
This research is part of the programme DesCartes and is supported by the National Research Foundation, Prime Minister’s Oﬃce,
Singapore under its Campus for Research Excellence and Technological Enterprise (CREATE) programme.

\subsection*{Competing interests}
The authors declare that they have no competing interests.

\subsection*{Ethics approval and consent to participate}
Not applicable

\subsection*{Data availability}
Data will be made available on reasonable request to the authors.

\end{document}